# Unveiling the factors of aesthetic preferences with explainable AI

Derya Soydaner | Johan Wagemans

Department of Brain and Cognition, University of Leuven (KU Leuven), Leuven, Belgium

**Correspondence**
Derya Soydaner, Department of Brain and Cognition, University of Leuven (KU Leuven), Leuven 3000, Belgium.
Email: derya.soydaner@kuleuven.be

**Funding information**
European Research Council, Grant/Award Number: 101053925

**Abstract**
The allure of aesthetic appeal in images captivates our senses, yet the underlying intricacies of aesthetic preferences remain elusive. In this study, we pioneer a novel perspective by utilizing several different machine learning (ML) models that focus on aesthetic attributes known to influence preferences. Our models process these attributes as inputs to predict the aesthetic scores of images. Moreover, to delve deeper and obtain interpretable explanations regarding the factors driving aesthetic preferences, we utilize the popular Explainable AI (XAI) technique known as SHapley Additive exPlanations (SHAP). Our methodology compares the performance of various ML models, including Random Forest, XGBoost, Support Vector Regression, and Multilayer Perceptron, in accurately predicting aesthetic scores, and consistently observing results in conjunction with SHAP. We conduct experiments on three image aesthetic benchmarks, namely Aesthetics with Attributes Database (AADB), Explainable Visual Aesthetics (EVA), and Personalized image Aesthetics database with Rich Attributes (PARA), providing insights into the roles of attributes and their interactions. Finally, our study presents ML models for aesthetics research, alongside the introduction of XAI. Our aim is to shed light on the complex nature of aesthetic preferences in images through ML and to provide a deeper understanding of the attributes that influence aesthetic judgements.

## BACKGROUND

Why do some images appeal to us while others evoke the opposite reaction? This question remains largely unanswered, as aesthetic preferences vary among individuals and depend on numerous factors. However, certain aesthetic attributes play a significant role in shaping these preferences. Insights from diverse fields, including psychology and computer science, have contributed valuable knowledge on aesthetic preferences. In general, psychological studies in the field of empirical aesthetics have primarily focused on exploring the factors influencing aesthetic preferences (Nadal & Vartanian, 2019). In computer science, the emphasis has been on treating image aesthetic assessment as an artificial intelligence (AI) task, leading to models for classifying images based on aesthetic qualities or making aesthetic predictions (Deng et al., 2017). The interdisciplinary field of computational aesthetics is a blend of these two disciplines, dedicated to developing computational methods for aesthetics research (Brachmann & Redies, 2017; Hoenig, 2005; Valenzise et al., 2022).

However, there is a wider range of models available than those currently utilized in computational aesthetics. For instance, in research aligned with our direction in the literature, Conwell et al. (2021) applied deep neural networks. Iigaya et al. (2021) explored the prediction of aesthetic preferences for visual art using a combination of different visual features. They developed a computational framework that includes a Linear Feature Summation (LFS) model and a Deep Convolutional Neural Network (DCNN) to assess how aesthetic values are formed. They also applied ridge regression to predict liking ratings. More recently, Iigaya et al. (2023) implemented LFS and DCNN again, this time using fMRI data and participants' ratings for paintings. Farzanfar and Walther (2023) applied random forest to predict aesthetic values for scenes, employing contour properties as features.

In this study, we present alternative models, adopting a machine learning (ML) approach to deepen our understanding of aesthetic preferences in images within the realm of computational aesthetics. In addition, to enhance the interpretability of the ML models, we employ the SHapley Additive exPlanations (SHAP) (Lundberg & Lee, 2017), a prominent Explainable AI (XAI) technique. SHAP is specifically designed to provide insights into the contributions of individual attributes to a model's output. It achieves this by assigning 'SHAP values' to each feature, indicating its contribution to the model's output. It clarifies the importance of each input in making predictions and also examines their interactions. This approach provides a deeper understanding of how specific features influence the model's predictions. In terms of data, we utilize benchmark datasets in aesthetics research. Consequently, our investigation results in a methods paper that analyses aesthetic benchmarks in the literature, supported by XAI. Our unique perspective focuses on understanding the influence of various attributes on aesthetic judgements. We acknowledge existing studies that have investigated the relationship between (subjective) object features and aesthetic evaluations (for extensive reviews, see Nadal & Vartanian, 2019). However, our paper takes a different focus. We analyse current image aesthetic benchmarks to demonstrate how they can be better understood, and we propose a new methodology. While many studies in the field of AI focus on image aesthetic assessment models that process image data as input to predict aesthetic-related scores as output (e.g., Celona et al., 2022; Li, Huang, et al., 2023; Li, Zhu, et al., 2023; Lu et al., 2014; Pan et al., 2019; Soydaner & Wagemans, 2023; Talebi & Milanfar, 2018), we take a novel approach by shifting our focus to aesthetic-related scores themselves. Instead of utilizing images as inputs, we develop regression models that take aesthetic attribute scores as inputs to predict the overall aesthetic scores of images. This alternative approach allows for a more detailed analysis of attribute information in aesthetic image datasets, providing valuable insights into the factors that contribute to aesthetic preferences.

While SHAP effectively highlights the importance of each attribute on the model's predictions, it does not assess prediction quality itself. To comprehensively evaluate our approach, we employ a diverse set of ML models. Specifically, we utilize two ensemble ML models, namely Random Forest and eXtreme Gradient Boosting (XGBoost), along with a kernel-based regression model known as Support Vector Regression (SVR). Additionally, we incorporate a neural network approach, specifically the Multilayer Perceptron, which is particularly well suited for regression problems. By employing multiple models and consistently observing results in conjunction with SHAP, we establish a reliable interpretation of the effects of image aesthetic attributes. Furthermore, we evaluate our approach on three image aesthetic benchmarks. Since only three benchmarks in the literature include aesthetic attribute information, we can utilize them to explore attributes and identify similarities across datasets.

We summarize the key contributions of this work as follows:

- We introduce a novel perspective by utilizing ML models for regression to gain insights into aesthetic preferences in images.
- We provide the first detailed comparative analysis of various ML models within the computational aesthetics field, exploring their performance in predicting aesthetic scores.
- To the best of our knowledge, we pioneer the utilization of attribute information in image aesthetic benchmarks through a data mining approach, providing a deeper analysis of the role of attributes in aesthetic judgements.
- We present the first application of the SHAP method in understanding image aesthetics, enhancing the interpretability of ML models, and unveiling the contributions of individual attributes to aesthetic predictions.

Before explaining the methodology of our study and reporting and discussing its main findings, we would like to point out why we think this is an interesting case study for the use of AI in psychology. Empirical aesthetics, with its focus on trying to address the factors underlying aesthetic preferences (Nadal & Chatterjee, 2019), could be regarded as a small, somewhat peripheral area of research in psychology because of its very specific goal. We would like to argue, however, that it is an excellent example of what psychology is about and what kind of challenges it faces. Psychology is aimed at providing causal explanations of human behaviour. By generalizing over many kinds of behaviours and explanations, psychology also hopes to provide overarching theories of the mind. By doing so, it faces two main problems: many of the explanatory variables concern hidden processes and a specific type of behaviour is always driven by a complex mixture of causal factors, which can differ strongly depending on the circumstances (experimental conditions as well as complex situations from everyday life). All of this is very obvious in empirical aesthetics as well. The preference (or appreciation) of a particular image or artwork results from a combination of processes in perception, cognition, and emotion (the so-called "central triad in psycho-aesthetics," (Nadal & Chatterjee, 2019)). Each of these is also influenced by social and cultural factors, as well as individual differences resulting from personality and expertise. This interplay, and its dynamic variations due to context variables, was nicely illustrated in the model by Leder and colleagues published in this journal 20 years ago (Leder et al., 2004, see also the updated version of 10 years ago Leder & Nadal, 2014). So, aesthetic preference offers an ideal playground to see how all of the main areas in psychology work together.

Regarding the methodological challenges, the problems of empirical aesthetics should also be clear: given the complex mixtures of causal factors and their contextual dependencies, it is almost impossible to generalize from specific results of specific studies (with small samples of stimuli and participants) to universal theories. The hidden nature of the psychological processes at play and the difficulties to manipulate the specific hypothesized subprocesses directly (let alone in isolation or parametrically) limit the possibilities of experimental research even further. It is here that computational aesthetics and ML could come to the rescue. By training AI models on extensive benchmark datasets, the hope is that we could infer the relative contribution of a number of factors (at least those that can be computed from data in images). The interplay between the factors can be captured better by more complicated models than linear regression. The transparency of most of these AI models can be improved by examining the contribution of different factors by using SHAP.

## METHODOLOGY

In this study, our objective is to present a diverse set of ML models in the field of empirical aesthetics and to gain insights into image aesthetic preferences through a data mining approach. To achieve this, we leverage the SHAP method, a popular XAI technique, to provide explanations for our models. Our methodology comprises two main steps: first, training an ML model that takes aesthetic attributes as inputs to predict the overall aesthetic scores of images, and subsequently employing the SHAP method to explain the importance of these inputs in predictions. Figure 1 illustrates the general overview of our approach.

We employ a diverse range of ML models, comparing their performances, and then examining the SHAP results of the model that achieves the most accurate predictions. This section provides a detailed explanation of the models we employ in this study, along with an in-depth description of how we leverage the SHAP method to gain insights into the attributes driving aesthetic preferences.

We begin by implementing ensemble methods that utilize decision trees, specifically Random Forest and XGBoost. Ensemble learning is a widely used technique in ML that combines multiple individual models to create a more accurate predictive model. The core idea behind ensemble learning is to combine predictions from diverse models, leveraging the strengths of each model while mitigating their weaknesses. The individual models within the ensemble can be of the same type, such as multiple decision trees, or they can be different types, such as a combination of decision trees, support vector machines (SVMs), and neural networks.

Continuing with our methodology, we incorporate a kernel-based approach, specifically SVR (Drucker et al., 1996), which leverages the powerful mathematical foundations of SVMs (Boser et al., 1992) to capture intricate relationships within the data. Additionally, we develop a Multilayer Perceptron, a neural network architecture known for its suitability in regression tasks. We provide a comprehensive analysis of these diverse ML models to gain insights into their predictive capabilities.

Through this analysis, our objective is to identify the model that demonstrates the highest performance in predicting image aesthetics. Once the best performing model is determined, we proceed with the utilization of the SHAP method to interpret the importance of attributes in predicting the overall aesthetic score. By applying the SHAP method, we aim to gain insights into the factors influencing image aesthetics and enhance the interpretability of the regression models.

## Machine learning models for regression

Our focus is predicting aesthetic scores of images, that is, a regression task. Therefore, the ML models discussed in this section are primarily utilized for regression purposes. As a reminder, the main distinction between regression and classification tasks lies in the nature of the output: regression tasks aim at predicting continuous values, while classification tasks are about determining the discrete class labels an input belongs to. It is important to note that ML models detailed in this section can also be applied to classification tasks. However, in our research, we specifically implement them within the context of regression.

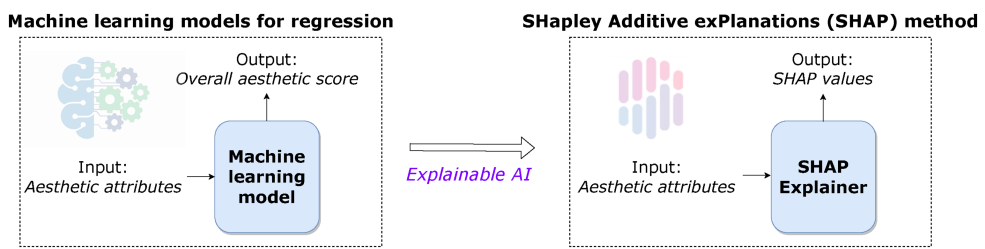

**FIGURE 1** Overview of our approach: Training a machine learning model using various aesthetic attribute scores to predict overall aesthetic score, and then employing a SHAP explainer based on the trained machine learning model to compute SHAP values for the same attributes.

## Random forest

Random Forest, also known as Random Decision Forest, was initially introduced by Ho (1995) as an ensemble learning method. It constructs multiple decision trees in randomly selected subspaces of the feature space. Decision trees are powerful models for classification and regression tasks; for more detailed information, see (Alpaydın, 2014; Géron, 2017).

The concept of Random Forest was further developed by Breiman (2001), building upon the principles of *bagging*. In bagging, the same training algorithm is applied to each predictor, but they are trained on different random subsets of the training set. When sampling is performed *with* replacement, this method is referred to as bagging (Breiman, 1996) in ML, or *bootstrapping* in statistics. Conversely, when sampling is performed *without* replacement, it is known as *pasting*. Random Forest is an ensemble of decision trees that are typically trained using the bagging method and occasionally with pasting as an alternative approach (Géron, 2017).

Random Forest constructs multiple decision trees, each trained on different random subsets of the training data. This results in a diverse set of predictors working collectively to make predictions. Once all predictors are trained, the ensemble can make a prediction for a new instance by computing mean or average prediction across the individual trees, particularly for regression tasks. Random Forest incorporates additional randomness during the tree-growing process. Instead of searching for the optimal feature when splitting a node, it selects the best feature among a random subset of features (Géron, 2017). This strategy often leads to a more robust and improved model.

## XGBoost

eXtreme Gradient Boosting, commonly known as XGBoost, is another ensemble learning method that combines the predictions of multiple models to enhance prediction accuracy (Chen & Guestrin, 2016). It is built upon the principles of *boosting* (Schapire, 2017), an ensemble technique widely used in ML. Boosting methods train predictors sequentially, with each subsequent model aiming to correct the errors of its predecessor. One popular boosting algorithm is *Gradient Boosting*, which integrates predictors incrementally into the ensemble. Each new predictor aims to reduce the *residual errors* generated by the preceding predictor (Géron, 2017). More specifically, each model tries to predict the residuals calculated by its predecessor, using a gradient descent algorithm to minimize the loss.

XGBoost is a scalable ML framework specifically designed for *tree* boosting (Chen & Guestrin, 2016). This algorithm constructs decision trees sequentially and iteratively adds them to the ensemble while optimizing a loss function. Each subsequent tree is built to correct the errors made by the preceding trees. XGBoost learns through gradient boosting, which updates the model's parameters using gradients to minimize the loss function. Therefore, XGBoost can be regarded as an implementation of gradient tree boosting (Friedman, 2001), effectively combining the strengths of decision trees and gradient-based optimization techniques.

XGBoost is known for its high predictive accuracy, scalability, and flexibility. It places significant emphasis on the use of weights, which are assigned to all independent variables, and fed into the decision tree during prediction process. Notably, XGBoost employs a strategy where the weights of variables that were incorrectly predicted by the tree are increased, and these adjusted variables are then passed on to the subsequent decision tree. This ensemble of individual predictors results in a strong and precise model. Additionally, it incorporates both $L^1$ and $L^2$ regularization techniques, providing effective control and penalization of the model. To elaborate, regularization is a technique used to prevent *overfitting*, which occurs when the model performs well on training data, yet poorly on test data. Goodfellow et al. (2016) provide a comprehensive exploration of regularization techniques.

## Support vector regression

SVR (Drucker et al., 1996) is a widely used ML method for regression tasks. It is rooted in the principles of SVMs (Boser et al., 1992), with the objective of finding a hyperplane that best fits the training data while minimizing the error. This hyperplane is defined by a set of *support vectors*, which are data points located closest to the boundary of the hyperplane. The distance between the support vectors and the hyperplane is known as the *margin*. In SVR, the aim is to position as many instances as possible on the hyperplane while controlling margin violations, which occur when instances fall outside the margin. The width of the hyperplane is controlled by the parameter $\varepsilon$ (Géron, 2017).

This approach brings the advantage of effectively capturing nonlinear dependencies, leading to improved performance (Schölkopf & Smola, 2002). To handle non-linear regression problems efficiently, SVR employs a key concept known as the *kernel trick*. The kernel trick becomes necessary when the input data cannot be linearly separated in its original feature space. It allows SVR to implicitly map the input data into a higher dimensional feature space, where it becomes linearly separable. This mapping is achieved using a kernel function, which computes the dot product between pairs of data points in the higher dimensional space without explicitly calculating the transformed feature vectors. One commonly used kernel function in SVR is the Radial Basis Function (RBF) kernel, defined as:

$$K(x, x') = \exp\left(-\frac{\|x - x'\|^2}{2\sigma^2}\right) = \exp\left(-\gamma \|x - x'\|^2\right) \quad (1)$$

where $x$ and $x'$ represent input data, $\|x - x'\|^2$ denotes the squared Euclidean distance between them, and $\gamma$ is the kernel coefficient that controls the smoothness of the kernel function. Equation 1 measures the similarity or dissimilarity between two data points based on their distance in the input feature space. It assigns higher similarity values to data points that are closer together and lower similarity values to those that are farther apart. By utilizing kernel functions, SVR can effectively find a linear hyperplane in this transformed space, enabling accurate regression predictions even when the original feature space lacks a linear relationship.

## Multilayer perceptron

Our final ML model for predicting aesthetic scores is a neural network, specifically a Multilayer Perceptron (MLP), which is well suited for regression tasks. It is a feedforward neural network with one or more hidden layers between the input and output layers, enabling it to extract meaningful features from the data.

The training process of an MLP involves two main steps: the forward pass and the backward pass. During the forward pass, the input is fed into the input layer, and each hidden layer computes a weighted sum of its inputs, followed by the application of an activation function. The activations then propagate forward through the network. An error is then computed using a loss function that compares predictions with ground-truth labels. In the backward pass, known as backpropagation (Rumelhart et al., 1986), the gradients of errors with respect to the network's weights are computed. These gradients are then used in the learning process, where an algorithm, usually called an *optimizer*, updates the weights to minimize the error (Goodfellow et al., 2016). For a more detailed explanation of MLP and its training process, see (Yüksel et al., 2021).

Neural networks, including the MLP, are powerful models in ML, particularly within the subfield of deep learning. They have gained immense popularity due to their ability to leverage multiple hidden layers and learn intricate representations from raw data. Deep learning has revolutionized various domains and led to significant advancements in AI, achieving breakthroughs in various fields such as natural language processing (Ouyang et al., 2022) and computer vision (Krizhevsky et al., 2012; Ramesh et al., 2021).

## Explainable AI

With the widespread adoption of AI techniques, there has been an increasing demand for explanations and transparency. ML models have often been criticized as 'black box' systems. This critique becomes particularly concerning when employing these models to gain insights about human intelligence and task performance (Bowers et al., 2023). Addressing this need, the field of XAI has gained significant attention (Biran & Cotton, 2017; Gohel et al., 2021; Holzinger et al., 2022; Linardatos et al., 2020). XAI can be used as a useful tool to augment the psychological insight gained from ML models, thereby enhancing the utility of AI as a valuable source in psychology.

Several XAI techniques have been proposed such as Local Interpretable Model-agnostic Explanations (LIME) (Ribeiro et al., 2016) and Deep Learning Important FeaTures (DeepLIFT) (Shrikumar et al., 2017). One prominent technique that has achieved widespread recognition is SHAP (Lundberg & Lee, 2017). SHAP values, the key component of this technique, prove to be more consistent with human intuition than other techniques that fail to meet three desirable properties for a single unique solution: local accuracy, missingness, and consistency (Lundberg & Lee, 2017). SHAP is a game-theoretic approach designed to explain the outputs of ML models, providing insights into the contribution of each feature to the model's predictions. By assigning SHAP values to features, which represent their relative importance compared to a baseline reference, this approach enables a comprehensive understanding of the factors that influence the model's output. Consequently, SHAP enhances the interpretability and transparency of AI systems.

The Shapley value, originally introduced by Shapley (Shapley, 1953), is a concept in Game Theory that assigns payouts to players based on their individual contributions to the total payout within a cooperative coalition. The concept of Shapley values has been extensively studied in Game Theory literature (Winter, 2002) and has emerged as a principled framework for obtaining feature attributions as explanations. In the context of XAI, the features of the model are treated as the players, while the prediction itself represents the 'game'. By employing the SHAP method, we aim to determine the extent to which each feature, or 'player' in the context of the coalition, contributes to the overall prediction.

The SHAP explanation method leverages the concept of Shapley values derived from coalitional game theory to quantify the contributions of individual features in an ML model. Shapley values provide a measure of the influence each feature has on the model's predictions, offering insights into how the 'payout' (i.e. the prediction) should be fairly distributed among the features (Molnar, 2022). Since the exact computation of SHAP values is challenging, Lundberg and Lee (2017) introduced model-type-specific approximation methods. Among these, *KernelSHAP* is designed for kernel-based models like SVR, while *TreeSHAP* is a highly efficient approach tailored for tree-based models such as Random Forest. These approximation methods enable the computation of SHAP values, facilitating the interpretation and explanation of predictions in various types of models. More information about SHAP values and approximation methods can be found in (Lundberg & Lee, 2017).

Despite the usefulness and widespread adoption of the SHAP method, one major challenge in utilizing Shapley values for model explanation is the significant computation time (Broeck et al., 2022), which grows exponentially with the number of features involved (Molnar, 2022). Fortunately, SHAP performs efficiently for tree-based ML models like XGBoost and Random Forest, as well as for the relatively simple MLP used in our study. However, it becomes notably slow for SVR. Managing the computational cost of SHAP becomes crucial, particularly when dealing with models involving a large number of features, and we will discuss potential mitigations and their implications in the subsequent sections.

The SHAP method offers versatility in visualizing and interpreting its results, making it applicable across various domains. While SHAP has been utilized in different areas, such as explaining image models (Lahiri et al., 2022), our study pioneers its application in understanding aesthetic preferences for images. By employing SHAP in this novel context, we aim to shed light on the underlying factors that contribute to image aesthetics and provide insights into the subjective nature of aesthetic judgements.

# DATASETS

In this study, we use three publicly available datasets specifically designed for image aesthetic assessment. These datasets include diverse attributes that are known to influence aesthetic preferences. Notably, each dataset not only provides overall aesthetic scores for images but also includes attribute scores, making them ideal for regression modelling. We provide a detailed description of each dataset, highlighting their unique characteristics and relevance to our research objectives. By leveraging these datasets, we aim to gain comprehensive insights into the factors influencing image aesthetics.

## Aesthetics with attributes database

We begin with the aesthetics with attributes database (AADB) (Kong et al., 2016), which serves as a widely recognized image aesthetic benchmark. This dataset comprises 10,000 RGB images sourced from the Flickr website, each with a size of 256 × 256 pixels. Each image in the AADB dataset is accompanied by an overall aesthetic score provided by five different raters. Kong et al. (2016) reported the average aesthetic scores provided by these raters for each image, which serve as the ground-truth scores. These scores range from 1 to 5, with 5 representing the most aesthetically pleasing score.

Moreover, the AADB dataset includes 11 attributes that professional photographers have identified as influential in aesthetic judgements. These attributes are balancing element, interesting content, colour harmony, shallow depth of field, good lighting, motion blur, object emphasis, rule of thirds, vivid colour, repetition, and symmetry. Notably, each image in the AADB dataset is associated with scores for each attribute. Raters indicated whether each attribute has a positive, negative, or null (zero) effect on the aesthetics of an image, except for repetition and symmetry, where only the presence or absence of the attribute is rated. The raters were not permitted to assign negative scores for repetition and symmetry.

The scores provided in the AADB dataset are presented in a normalized form. The average aesthetic scores are normalized within the range of [0, 1], while all attributes, excluding repetition and symmetry, are normalized within the range of [−1, 1]. Repetition and symmetry scores, on the other hand, are normalized within the range of [0, 1]. Figure 2 illustrates two sample images from the training set of the AADB dataset, showcasing examples of both low and high aesthetics. The figure includes the corresponding overall aesthetic scores as well as attribute scores.

The AADB dataset has been officially partitioned into three subsets, as described by Kong et al. (2016). Specifically, the dataset is divided into 500 images for validation, 1000 images for testing, and the remaining images for training. In our experiments, we use this official partition to train and evaluate our ML models.

## Explainable visual aesthetics dataset

The Explainable Visual Aesthetics (EVA) dataset (Kang et al., 2020) is a representative collection of 4070 images, each of which has been rated by a minimum of 30 participants. Within the EVA dataset, each image receives between 30 and 40 votes. These ratings are assigned on an 11-point discrete scale, with the extremes of the scale labelled as 'least beautiful' (corresponding to 0) and 'most beautiful' (corresponding to 10). Alongside the aesthetic ratings, the EVA dataset includes four attributes: light and colour, composition and depth, quality, and semantics of the image. Participants provided ratings for each attribute on a four-level Likert scale, ranging from 'very bad' to 'bad', 'good', and 'very good'. Figure 3 presents two sample images from the EVA dataset, illustrating examples of both low and high aesthetic images.

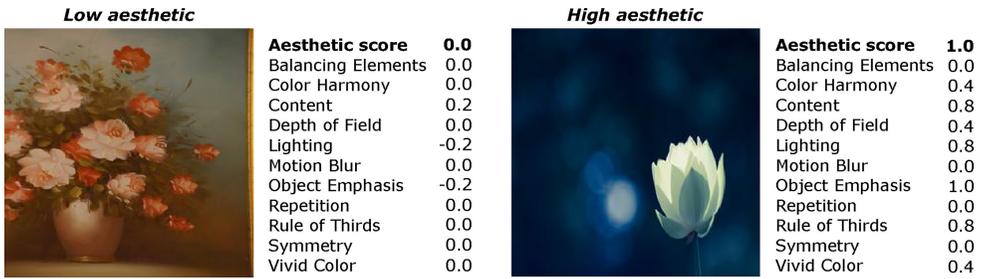

**FIGURE 2** Example images from the training set of the AADB dataset. Each image has an overall aesthetic score and scores for 11 attributes. (Left) Low aesthetic: An image rated with the lowest overall aesthetic score. (Right) High aesthetic: An image rated with the highest overall aesthetic score. It is important to note that the AADB dataset includes multiple images with overall aesthetic scores of 0.0 and 1.0. We randomly selected two examples for illustration.

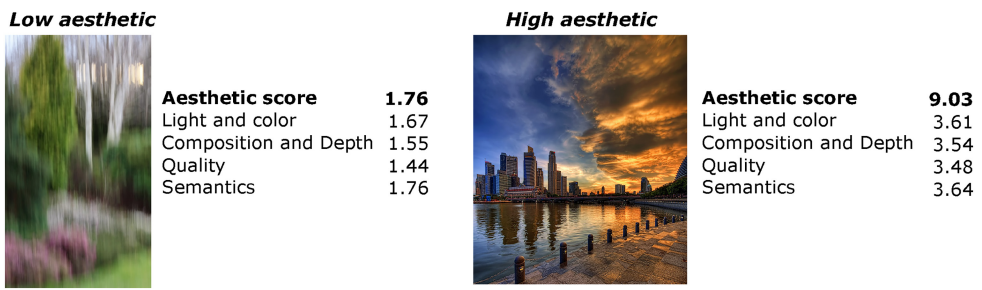

**FIGURE 3** Example images from the training set of the EVA dataset. Each image has an overall aesthetic score and scores for four attributes. (Left) Low aesthetic: The image rated with the lowest overall aesthetic score. (Right) High aesthetic: The image rated with the highest overall aesthetic score.

In contrast to the AADB dataset, Kang et al. (2020) provided the complete set of ratings from the participants for each image in the EVA dataset. In order to facilitate our analysis, we calculated the average scores based on these ratings. Unlike the AADB dataset, which offers predefined train–validation–test splits, the EVA dataset does not have an official partition. This is because Kang et al. (2020) did not employ any specific prediction model. As a result, we align our approach with previous studies that focus on image aesthetic assessment using the EVA dataset. In the literature, different training and testing splits have been applied for the EVA dataset (Li, Zhi, et al., 2023; Shaham et al., 2021). For our experiments, we follow the three studies (Duan et al., 2022; Li, Huang, et al., 2023; Soydaner & Wagemans, 2023), which use 3500 images for training and 570 for testing.

## Personalized image aesthetics database with rich attributes

The final dataset utilized in this study is the Personalized image Aesthetics database with Rich Attributes (PARA) (Yang et al., 2022), which consists of 31,220 images collected from CC search.[1] To ensure a diverse range of content, the authors employed a pre-trained scene classification model to predict scene labels for the images. Subsequently, approximately 28,000 images were selectively sampled based on these predicted labels.

---

[1] https://search.creativecommons.org/.

**TABLE 1** Image aesthetic benchmarks and their corresponding attributes used in this study.

| Datasets | Attributes |
| --- | --- |
| AADB Kong et al. (2016) | Balancing elements, colour harmony, content, depth of field, light, motion blur, object emphasis, repetition, rule of thirds, symmetry, vivid colour |
| EVA Kang et al. (2020) | Light and colour, composition and depth, quality, semantics |
| PARA Yang et al. (2022) | Quality, composition, colour, depth of field, light, content, object emphasis |

To further refine the aesthetics score distribution, the PARA dataset was augmented with around 3000 additional images. These additional images were selected to provide clear aesthetics ground truth and were sourced from Unsplash,[2] as well as image quality assessment databases such as SPAQ (Fang et al., 2020) and KonIQ-10K (Hosu et al., 2020). The aim of this augmentation process was to achieve a balanced representation of aesthetics scores across the dataset, ensuring a comprehensive and diverse collection of images suitable for analysis.

Each image in the PARA dataset has been annotated by 25 subjects on average, with a total of 438 subjects contributing to the annotations. Each image is annotated with four human-oriented subjective attributes (emotion, difficulty of judgement, content preference, and willingness to share) and nine image-oriented objective attributes (aesthetics, quality, composition, colour, depth of field, content, light, object emphasis, and scene categories). In our study, we aim to predict the aesthetics scores using the image-oriented objective attributes (excluding the scene categories) as inputs for our regression models. We do not include the human-oriented subjective attributes and scene categories as the inputs since they are considered less relevant for our specific research objective. By utilizing these seven inputs, our ML models predict the aesthetic scores of the images. A summary of the attributes used as inputs for the ML models in this study, including those from the PARA dataset and other datasets, is provided in Table 1.

The image-oriented attributes in the PARA dataset are mostly discretely annotated on a scale from 1 to 5. Specially, the object emphasis attribute is represented by a binary label, indicating the presence or absence of a salient object within the image. The aesthetics score is assigned as a discrete class label, ranging from 1 to 5, reflecting the comprehensive judgement of visual aesthetics. To address ambiguity, Yang et al. (2022) introduced an additional option between each integer value. The quality score represents the overall judgement of image quality, also ranging from 1 to 5. A higher quality score denotes a better perceptual quality. It is worth mentioning that images with low perceptual quality in the PARA dataset exhibit various degradations, such as motion blur, JPEG compression, and others. Figure 4 presents two sample images from the PARA dataset, illustrating examples of both low and high aesthetic images.

## IMPLEMENTATION DETAILS

We conducted a systematic experimentation process, comparing the performance of four regression models: Random Forest, XGBoost, SVR, and MLP. Our goal is to find the model that achieves the highest performance, which we subsequently analyse in detail using the SHAP method. Each regression model is trained on three separate datasets, as outlined in Section Datasets. We employ appropriate hyperparameter configurations for each model, which are provided below. These specific hyperparameter settings are chosen based on empirical evidence and commonly accepted practices in the field of ML. Subsequently, we evaluate each trained model based on the performance metrics described in Section Experimental results.

---

[2]Unsplash, https://unsplash.com/.

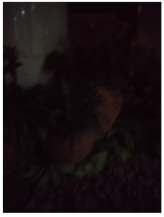 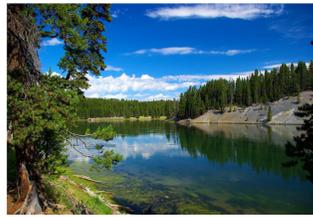

**FIGURE 4** Example images from the training set of the PARA dataset. Each image has an overall aesthetic score and scores for seven attributes. (Left) Low aesthetic: The image rated with the lowest overall aesthetic score. (Right) High aesthetic: The image rated with the highest overall aesthetic score.

## Model hyperparameter details for random forest

We outline the main hyperparameters and their corresponding values for the Random Forest model in our experiments. The first hyperparameter, *number of estimators*, determines the number of decision trees within the random forest. In our implementation, we set this parameter to 150, indicating the construction of 150 decision trees. Bootstrap samples, known as bagging in ML literature, are utilized during the tree-building process (see Section Random forest). The *criterion* parameter is a crucial factor in determining the split quality at each node within the decision trees. In our study, we utilize the mean squared error as the criterion to guide the splitting process.

The *maximum depth* parameter controls the depth of the decision trees. By default, when set to 'None', the trees continue growing until all leaves are pure or until the number of samples within a leaf falls below the *minimum samples split* threshold. This threshold determines the minimum number of samples required to initiate a split at an internal node. With a default value of 2, a node will only be split if it contains at least two samples. In our experiments, we do not explicitly set a maximum depth, allowing the trees to grow until these conditions are met. Similarly, the *minimum samples leaf* sets the minimum number of samples required for a node to be considered a leaf. In our implementation, we utilize a default value of 1, indicating that even the smallest node is eligible to be classified as a leaf. Additionally, the *maximum features* control the number of features considered when searching for the best split. In our experiments, we set the default value to 'auto', which selects the square root of the total number of features for consideration during the split selection process.

## Model hyperparameter details for XGBoost

We present the specific hyperparameter details for XGBoost used in our experiments. The *number of estimators*, corresponding to the number of boosting rounds or decision trees constructed, is set to 150 in our implementation. Each decision tree within the ensemble has a *maximum depth* of 3, limiting the complexity and depth of the individual trees. The *learning rate* is applied to each boosting iteration is set to 0.1. This parameter controls the contribution of each tree within the ensemble, striking a balance between the learning speed and the model's ability to generalize.

In our experiments, we use a *subsample* value of 1, indicating that the entire training dataset is used for constructing each tree within the ensemble. To introduce regularization and prevent overfitting, we incorporate the $L^1$ and $L^2$ regularization terms on the weights, with values of 0 and 1, respectively. The mean squared error is employed as the loss function to be minimized during training, while the root mean squared error served as the evaluation metric for validation during the training process.

## Model hyperparameter details for support vector regression

We provide an overview of the specific hyperparameters and their settings for SVR employed in our experiments. The choice of kernel function is crucial for mapping the input data to a higher dimensional feature space. We utilized the radial basis function (RBF) kernel, which is a popular choice for SVR. The regularization parameter, denoted as $C$, balances the trade-off between maximizing the margin and minimizing the training error. A smaller $C$ value results in a wider margin but may lead to more margin violations (Géron, 2017). In our experiments, we set $C$ to 1.0, striking a balance between margin width and training error. The $\varepsilon$ parameter defines the margin of tolerance within which no penalty is given to errors in the epsilon-insensitive loss function. We set $\varepsilon$ to 0.01, determining the acceptable range where errors are not penalized.

The kernel coefficient ($\gamma$) determines the influence of each training example, with higher values of $\gamma$ result in closer training examples having a stronger influence. We set $\gamma$ to 'scale', which automatically calculates an appropriate value based on the inverse of the feature scale. The *tolerance* for the stopping criterion, indicating the desired precision for convergence, is set to 1e-3. Lastly, the maximum number of iterations to perform is set to −1, indicating no explicit limit on the number of iterations. The algorithm continues until the convergence criteria are met.

## Model hyperparameter details for multilayer perceptron

In this study, our MLP architecture consists of one hidden layer with 32 hidden units. We determined this architecture through an iterative process, experimenting with different depths and numbers of hidden units. We found that using one hidden layer with 32 units is sufficient for achieving good performance while avoiding overfitting. Since we predict the aesthetic scores, which is a single numerical value, the output layer includes 1 unit. The hidden layer employs the Rectified Linear Unit (ReLU) (Glorot et al., 2011) activation function, while the output layer applies a linear function. ReLU activation outputs the input itself if it is positive; otherwise, it outputs zero. Linear activation in the output layer is often used in regression tasks where the goal is to predict a continuous value, and there is no need to map the output to a specific range or set of classes. We initialize the layer weights using the Glorot normal initializer, also known as Xavier normal initializer (Glorot & Bengio, 2010). We apply $L^2$ regularization to the hidden layer. The regularization term is set to 1e-2. We train our MLP architecture for 10 epochs using Adam algorithm (Kingma & Ba, 2014) to minimize mean squared error. The minibatch size is set to 64. The Adam algorithm is an adaptive gradient method that individually adapts the learning rates of model parameters (Kingma & Ba, 2014). During training, the Adam algorithm calculates the estimates of the first and second moments of the gradients and then utilizes decay constants to update them. These decay constants are additional hyperparameters along with the learning rate. More detailed information about the adaptive gradient methods in deep learning can be found in (Soydaner, 2020). In our study, the initial learning rate is 0.001, and decay constants are 0.9 and 0.999, respectively.

# EXPERIMENTAL RESULTS

In this section, we present the experimental results, following the methodology outlined in Figure 1. Initially, we develop individual ML models for each dataset to assess their performances in predicting overall aesthetic scores. For this evaluation, we report standard evaluation metrics, including the coefficient of determination ($R^2$), mean absolute error (MAE), mean squared error (MSE), and root mean squared error (RMSE). Additionally, we report Spearman's rank correlation ($\rho$), which is significant at $p < .01$ in our analysis. These metrics serve as reliable indicators of the accuracy and goodness of fit in our models.

After evaluating the models, we identify the one that demonstrates the highest predictive capability among them. Subsequently, we conduct a detailed examination of the results from this model using the SHAP method. By utilizing this XAI technique, we aim to gain a better understanding of the importance of individual attributes in predicting overall aesthetics scores.

## Analysis results on the AADB dataset

We begin our analysis with the AADB dataset (Kong et al., 2016), which is the oldest among the three image aesthetic assessment datasets described in Section Datasets. For our regression task, we train four different ML models described in Section Machine learning models for regression. Using the attribute scores as inputs (Table 1), these models predict the overall aesthetic scores of the images. Performance comparison of these models is based on various metrics, including the $R^2$ coefficient, MAE, MSE, RMSE, and $\rho$ between the predicted overall aesthetic scores and the ground-truth scores, presented in Table 2. In general, all ML models demonstrate good performance on the AADB dataset. However, the XGBoost and SVR models perform slightly better than the others, with SVR demonstrating a slightly superior performance compared to XGBoost.

Next, we apply the SHAP method to analyse the results of the SVR model on the AADB dataset. We calculate SHAP values for the test data, with each SHAP value representing the contribution of an individual feature to the model's output. We present a summary of the SHAP values in Figure 5. This SHAP summary plot provides a visual illustration of the impact of each feature on the predictions made by the SVR model. The features are listed on the y-axis of the plot, ranked by their importance, with the most influential feature at the top and the least influential at the bottom.

According to the results, the attribute 'content' emerges as the most influential feature in predicting the overall aesthetic scores. It is followed by 'object emphasis' and 'colour harmony', both of which also make significant contributions to the predictions. It is worth noting that these top three inputs include two high-level attributes and one mid-level attribute. On the other hand, 'repetition', 'motion blur', and 'symmetry' are observed to have minimal impact on predicting the overall aesthetic scores, as evidenced by their placement at the bottom of the plot. It is important to consider that in the AADB dataset, these three attributes are predominantly rated as neutral, which might explain their lower impact. Further insights into the distribution of attributes in the AADB dataset can be found in Appendix A: Figure A1.

Concerning previous findings in empirical aesthetics, our study acknowledges a limitation: the current benchmarks do not adequately reflect features commonly studied in empirical aesthetics. For example, in the AADB dataset, the majority of symmetry ratings are neutral, resulting in a poor representation of symmetry. Consequently, in our results, symmetry appears to have minimal effect on the aesthetic scores. This finding is in contrast to established literature, which indicates that symmetry significantly impacts aesthetic appreciation. However, this discrepancy arises from the limitations of the dataset rather than the actual influence of the feature.

**TABLE 2** Performance of four machine learning models on the AADB dataset. We calculate all metrics on the test data to evaluate generalization ability.

| Model | $R^2$ | MAE | MSE | RMSE | $\rho$ |
|---|---|---|---|---|---|
| Random forest | .8373 | 0.0660 | 0.0067 | 0.0820 | .923 |
| Multilayer perceptron | .8426 | 0.0638 | 0.0065 | 0.0806 | .938 |
| XGBoost | .8631 | 0.0597 | 0.0056 | 0.0752 | .935 |
| Support vector regression | **.8739** | 0.0576 | 0.0052 | 0.0722 | .939 |

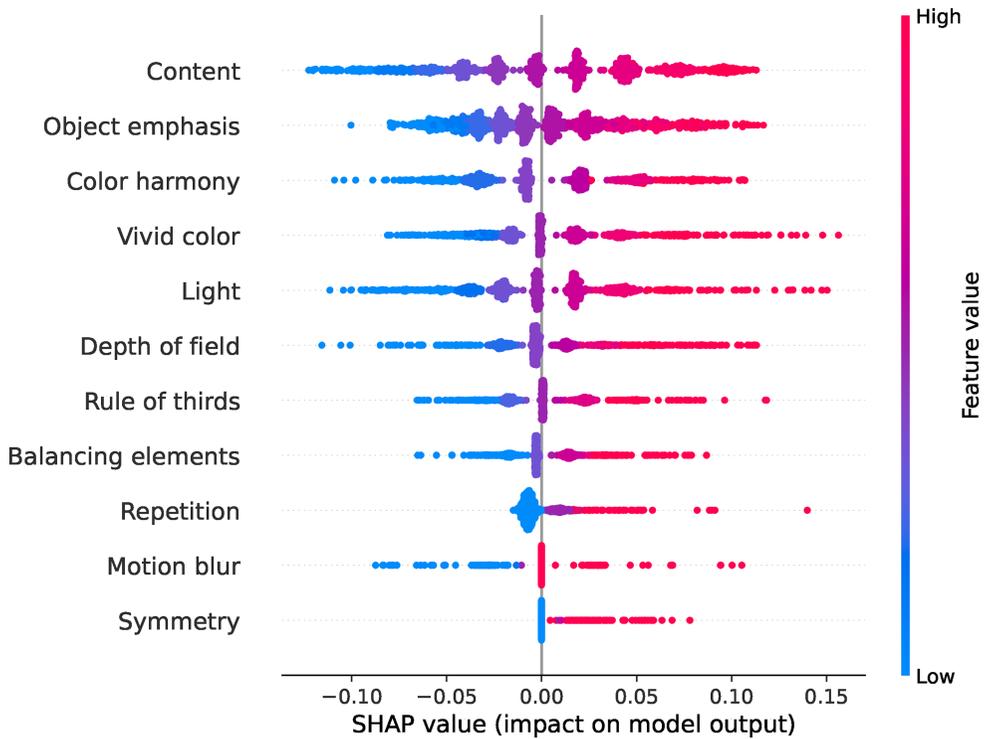

**FIGURE 5** SHAP summary plot for the AADB dataset based on the SVR model. Each dot represents a SHAP value for a feature across different observations, illustrating the feature's impact on individual predictions. The x-axis measures the magnitude of the SHAP values, indicating the relative importance of each feature: features with larger absolute SHAP values have a more substantial impact on the model's predictions, while those with smaller absolute SHAP values have a relatively minor impact. Each attribute is represented by a horizontal bar, with the length and position of each dot reflecting the SHAP value's magnitude and its positive or negative effect on the prediction. The colour represents the actual feature value for the corresponding observation, representing the direction of the feature's impact on the output: blue indicates a lower feature value, while red indicates a higher feature value. The colour intensity reflects the magnitude of the feature value, adding additional information about how the feature value correlates with its impact on the prediction. This colour coding helps to visually demonstrate how the actual feature value correlates with its impact on the model's output.

Due to the significant computation time required to compute SHAP values for the SVR model, we followed the recommendation in the official SHAP documentation[3] and applied k-means clustering to the training set. We summarized the training set with three clusters using k-means, with each cluster weighted by the number of points it represents. We also adopt the same k-means approach to examine the SHAP summary plot for the MLP model. For the other models (Random Forest and XGBoost), we directly examine the SHAP summary plots using the entire dataset without applying k-means. Remarkably, all these ML models consistently yielded similar results in terms of attribute rankings. In every case, the 'content' attribute emerged as the most influential in predicting the overall aesthetic score across all models. On the other hand, the 'repetition', 'motion blur', and 'symmetry' attributes consistently appeared at the bottom of the plot, indicating their minimal impact on the model's predictions. Across all four models, 'colour harmony' and 'object emphasis' consistently rank within the top three positions, alternating between second and third place. We can interpret this result as these two attributes having a similar level of importance in influencing the model's predictions. The ranking of the other attributes varied slightly across the models, but there are no striking changes in the overall results. For SHAP summary plots of the other ML models (Random Forest, XGBoost, and MLP), please refer to Appendix B: Figure B1.

---

[3]https://shap-lrjball.readthedocs.io/en/latest/examples.html.

SHAP values not only provide valuable insights into the individual contributions of features but also enable the examination of interactions between features. Interaction plots illustrate how two features jointly influence the model's prediction. Similar to the SHAP summary plot, in an interaction plot, a more intense red colour indicates higher feature values, while a more intense blue colour indicates lower feature values. By examining the interaction plots and observing the colour patterns, we can gain insights into how two features interact and jointly influence the model's predictions. This helps to identify important feature combinations and understand how the model makes decisions based on their joint values. In the AADB dataset, which comprises a total of 11 attributes, there are many potential interactions between these attributes. In Figure 6, we present some striking examples of the interactions observed in the SHAP values between aesthetic attributes. For example, we observe more positive interactions between balancing elements and content, as well as between colour harmony and content. These interactions suggest that when both of these attributes exhibit high values together, they tend to have a positive influence on the model's prediction, and their combined effect reinforces the model's output. Conversely, we observe the opposite effect with depth of field and object emphasis.

On the other hand, we see extremely positive interactions between attributes like balancing elements and motion blur, colour harmony and motion blur, content and motion blur, depth of field and motion blur, and light and motion blur. Motion blur appears to have a significant interaction with several other attributes, implying its importance when combined with these features in determining the overall aesthetic score. We present the rest of the interaction plots in Appendix C: Figures C1–C3.

## Analysis results on the EVA dataset

We continue our analysis with the EVA dataset. This time, we utilize four attribute scores as inputs for our ML models to predict the overall aesthetic scores of images. The results are presented in Table 3, where all ML models exhibit similar performances. Once again, the SVR model demonstrates slightly better performance, which leads us to apply the SHAP method to it for deeper insights into the importance of individual attributes in predicting overall aesthetic scores.

We present the SHAP summary plot for the EVA dataset in Figure 7. Here, the attribute 'semantics' emerges as the most important feature in predicting the overall aesthetic scores. It is followed by 'composition and depth', 'light and color', and 'quality', in that order. Interestingly, we observe some parallels between the results obtained from the EVA and AADB datasets. In the AADB dataset, colour harmony, vivid colour, and light are ranked as the third, fourth, and fifth most important attributes, respectively, while in the EVA dataset, light and colour emerges as the third most important attribute in predicting the overall aesthetic score. It is worth noting that each image in the EVA dataset has more ratings compared to the AADB dataset.

In the SHAP summary plots based on our Random Forest, XGBoost, and MLP models, the 'semantics' attribute is consistently ranked first, except in the Random Forest model. Additionally, 'quality' is consistently ranked last across all three models. Meanwhile, 'light and color' invariably appears in the third place. This aligns with findings in psychology literature that demonstrate how composition and semantics influence aesthetic preferences (Palmer et al., 2013). For SHAP summary plots of these ML models, please refer to Appendix D: Figure D1.

The interaction plots for attributes in the EVA dataset are presented in Figure 8. Since the EVA dataset has only four attributes as inputs, we have a smaller set of plots to examine compared to the AADB dataset discussed in the previous section. In all the interaction plots for the EVA dataset, we observe that high values of each feature combination tend to increase the model's output. Specifically, when the 'light and color' score is around 5, its interaction with 'composition and depth', 'quality', and 'semantics' strengthens their combined effect in a positive direction, respectively. Similarly, when the 'composition and depth' score is around 6, it exhibits interactions with 'quality' and 'semantics' that enhance their combined effect in a positive direction.

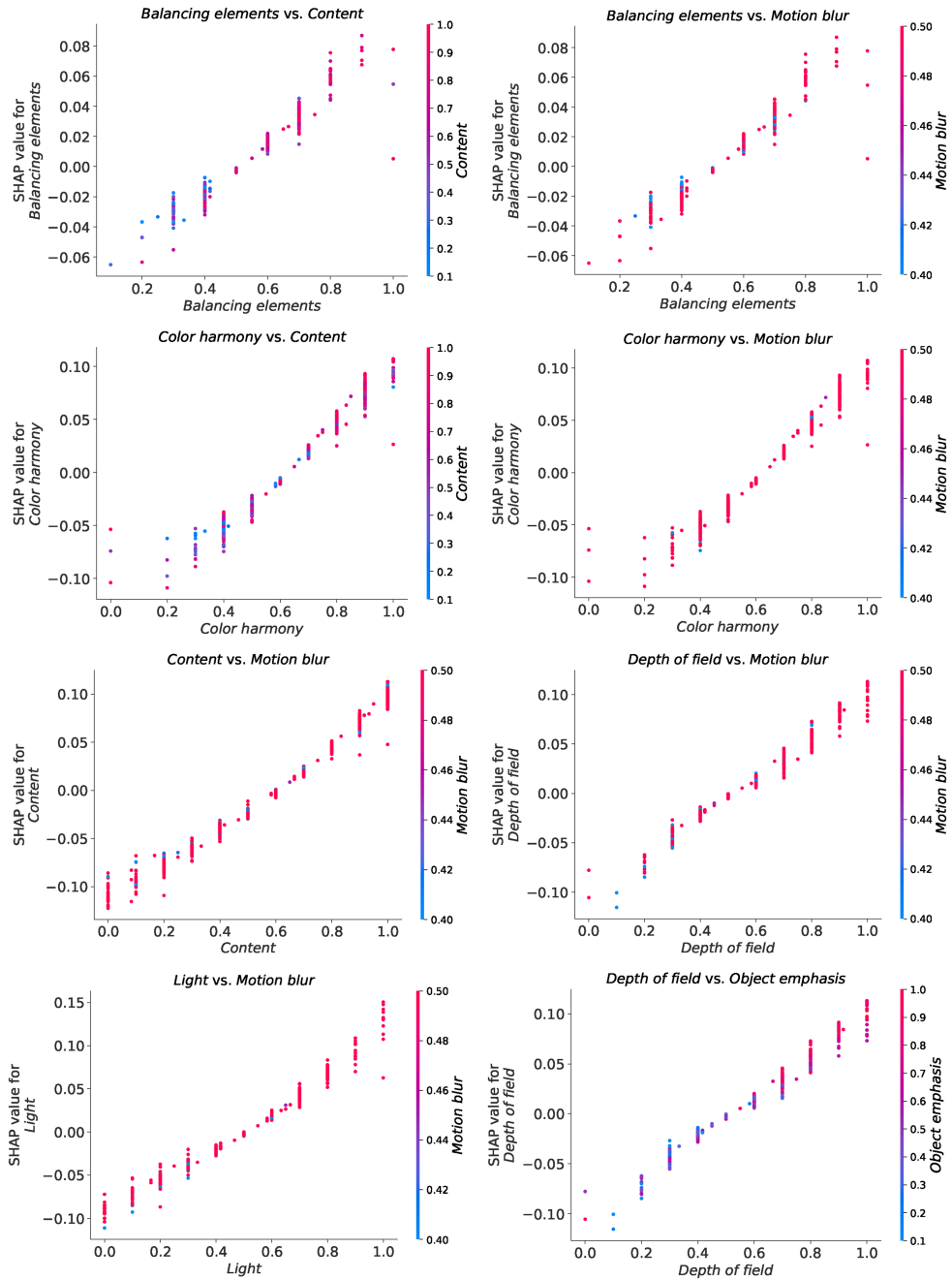

**FIGURE 6** Selected attribute interaction plots for the AADB dataset based on the SVR model. When the interaction plot shows more red, it suggests that both features together positively contribute to the model's prediction. High values of both features together tend to increase the model's output, and the interaction between these features strengthens their combined effect in a positive direction. Conversely, when the interaction plot exhibits more blue, it indicates that both features together contribute negatively to the model's prediction. Low values of both features together tend to decrease the model's output, and the interaction between these features strengthens their combined effect in a negative direction. The intensity of the colours reflects the strength of the interaction between the two features, indicating the magnitude of their combined impact on the model's prediction.

**TABLE 3** Performance of four machine learning models on the EVA dataset. We calculate all metrics on the test data to evaluate the generalization ability.

| Model | $R^2$ | MAE | MSE | RMSE | $\rho$ |
| --- | --- | --- | --- | --- | --- |
| Random forest | .9213 | 0.2272 | 0.0804 | 0.2836 | .959 |
| Multilayer perceptron | .9214 | 0.2262 | 0.0804 | 0.2835 | .959 |
| XGBoost | .9264 | 0.2172 | 0.0752 | 0.2743 | .961 |
| Support vector regression | **.9268** | 0.2174 | 0.0749 | 0.2737 | .962 |

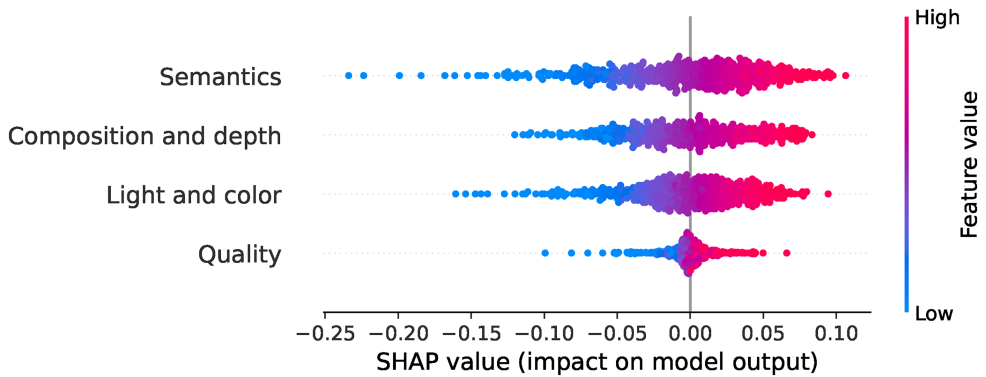

**FIGURE 7** Shap summary plot for the EVA dataset based on the SVR model.

Interestingly, the interaction between 'quality' and 'semantics' differs from the others. High-quality scores interact more positively with 'semantics', and both features contribute positively to the model's prediction, particularly when the quality score is around 8. Conversely, for values below than those thresholds, most attribute combinations contribute negatively to the model's prediction.

## Analysis results on the PARA dataset

The final dataset we use in our experiments is the PARA dataset (Yang et al., 2022). As shown in Table 1, we utilize seven attribute scores as inputs for the ML models to predict the overall aesthetic scores of images. The results are presented in Table 4. Although all models exhibit similar performance, the SVR model once again demonstrates slightly better results. Hence, we apply the SHAP method to the SVR model to observe the importance of individual attributes in predicting overall aesthetic scores.

We present the SHAP summary plot for the PARA dataset in Figure 9. Here, the attribute 'quality' emerges as the most important feature in predicting the overall aesthetic scores. We also know from psychology literature that high-quality images are often associated with positive aesthetic judgements (Tinio et al., 2011). It is followed by 'content', 'composition', 'colour', 'light', 'depth of field', and 'object emphasis', respectively. Interestingly, quality appears at the bottom of the SHAP summary plot for the EVA dataset in Figure 7, but it holds the top position in the SHAP summary plot for the PARA dataset. The other three ML models also yield the same result for 'quality' in their SHAP summary plots, as seen in the Appendix E: Figure E1. In the PARA dataset, 'quality' represents the overall judgement of image quality (Yang et al., 2022). In the EVA dataset, the question for each attribute was phrased as 'How do you like this attribute?' (Kang et al., 2020). The difference in the importance of 'quality' in the two datasets may be caused by the way the question was asked or the subjective nature of the attribute.

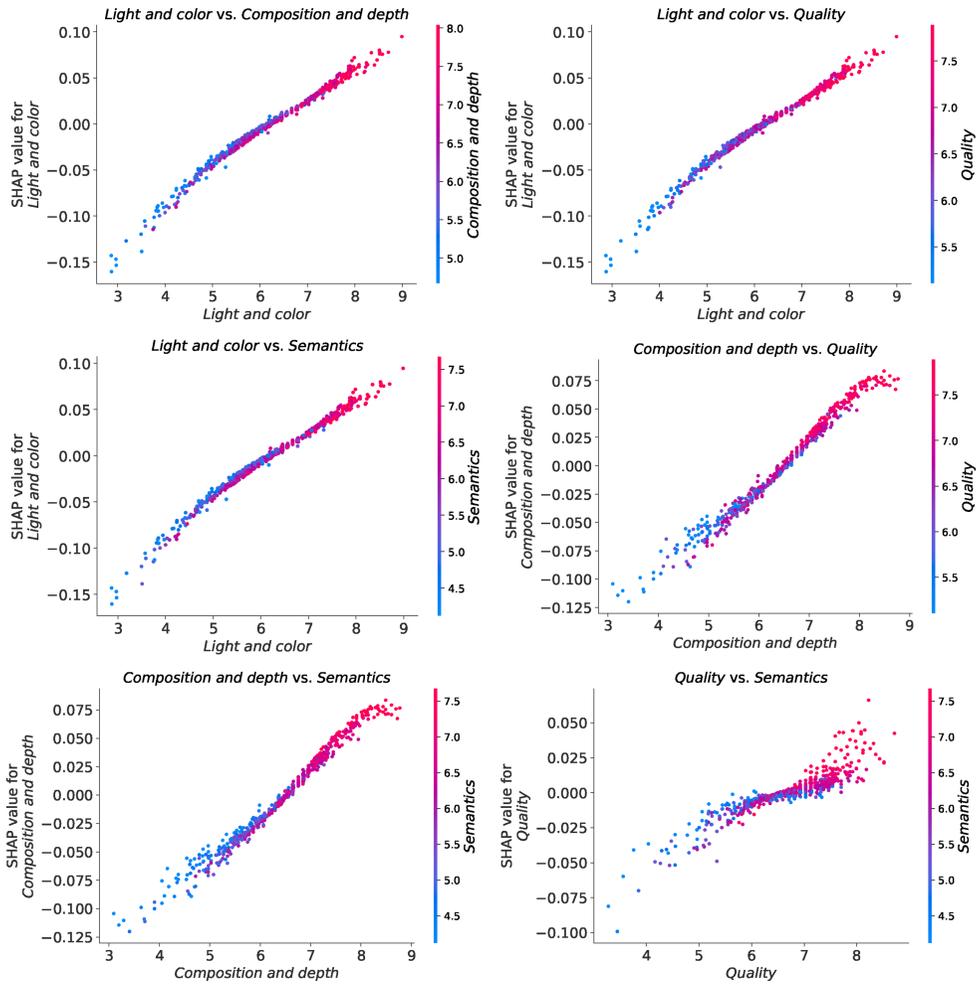

**FIGURE 8** Shap interaction plots for the EVA dataset based on the SVR model.

**TABLE 4** Performance of four machine learning models on the PARA dataset. We calculate all metrics on the test data to evaluate the generalization ability.

| Model | $R^2$ | MAE | MSE | RMSE | $\rho$ |
| --- | --- | --- | --- | --- | --- |
| Random forest | .9853 | 0.0510 | 0.0041 | 0.0644 | .988 |
| Multilayer perceptron | .9830 | 0.0554 | 0.0048 | 0.0694 | .987 |
| XGBoost | .9859 | 0.0502 | 0.0039 | 0.0631 | .989 |
| Support vector regression | **.9863** | 0.0495 | 0.0038 | 0.0623 | .989 |

In the PARA dataset, which includes a total of seven attributes, several interesting interactions between these attributes come to light. Figure 10 showcases some notable examples of these interactions as observed in the SHAP values between aesthetic attributes. The rest of the interaction plots are available in the Appendix F: Figure F1. One notable observation is that low values of the combination of 'quality' and 'object emphasis' tend to increase the model's output. This pattern holds consistent for each attribute combination with 'object emphasis'. Apart from this, we observe positive interactions between the other attributes. For example, when the quality score is higher than 3.5, it interacts with composition, reinforcing their combined effect in a positive direction. A similar pattern is observed for the attribute 'light' when paired with 'content'. On the other hand, when the depth of field score falls below 3, both this attribute and light contribute negatively to the model's prediction.

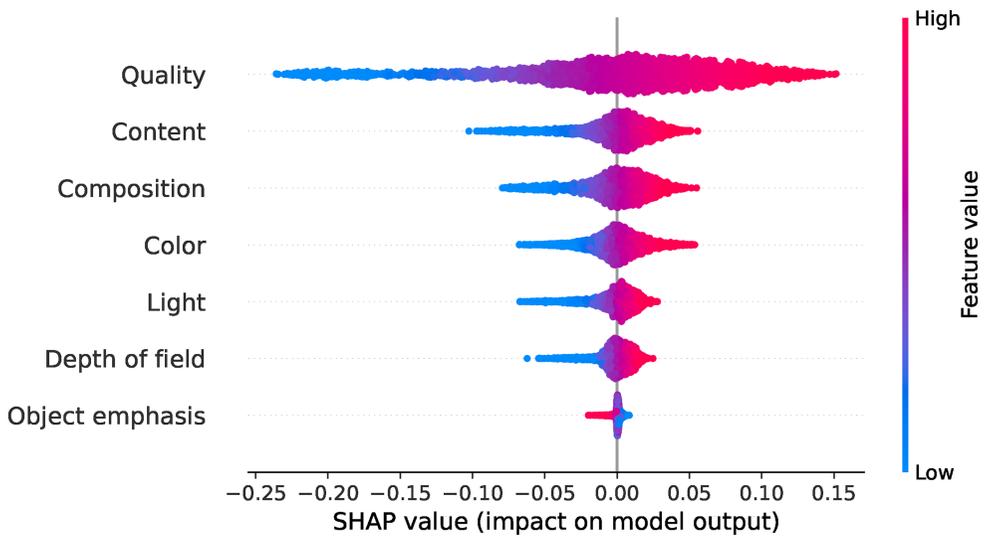

**FIGURE 9**  Shap summary plot for the PARA dataset based on the SVR model.

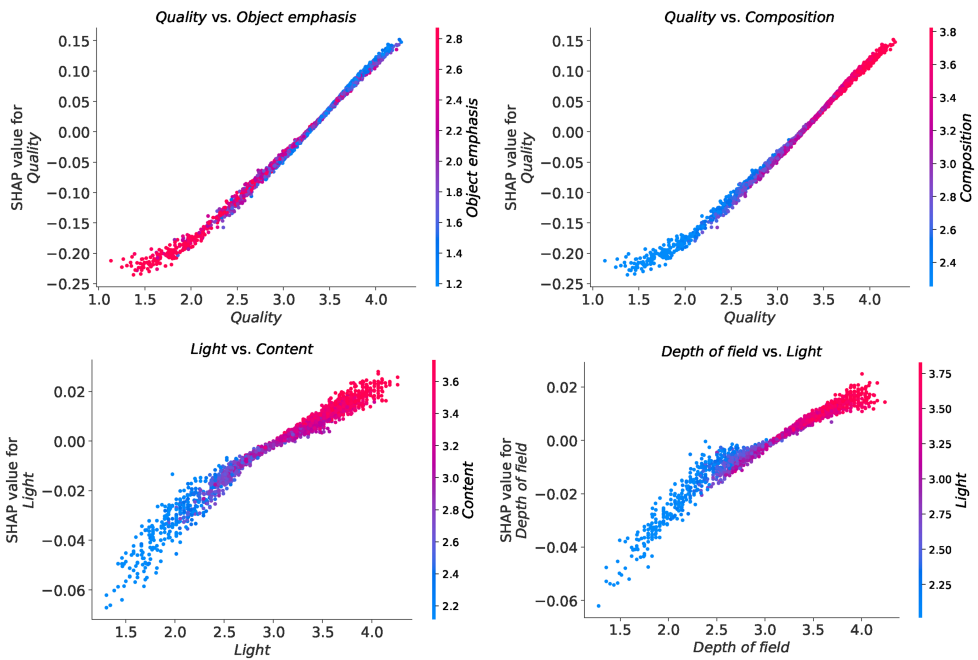

**FIGURE 10**  Selected attribute interaction plots for the PARA dataset based on the SVR model.

**TABLE 5** Performance comparison of the LR and SVR models on the three datasets. We calculate all metrics on the test data to evaluate the generalization ability.

| | $R^2$ | | MAE | | MSE | | RMSE | | $\rho$ | |
|---|---|---|---|---|---|---|---|---|---|---|
| Model | LR | SVR | LR | SVR | LR | SVR | LR | SVR | LR | SVR |
| AADB | .8653 | .8739 | 0.0588 | 0.0576 | 0.0055 | 0.0052 | 0.0746 | 0.0722 | .940 | .939 |
| EVA | .9277 | .9268 | 0.2174 | 0.2174 | 0.0739 | 0.0749 | 0.2720 | 0.2737 | .962 | .962 |
| PARA | .9860 | .9863 | 0.0500 | 0.0495 | 0.0039 | 0.0038 | 0.0629 | 0.0623 | .989 | .989 |

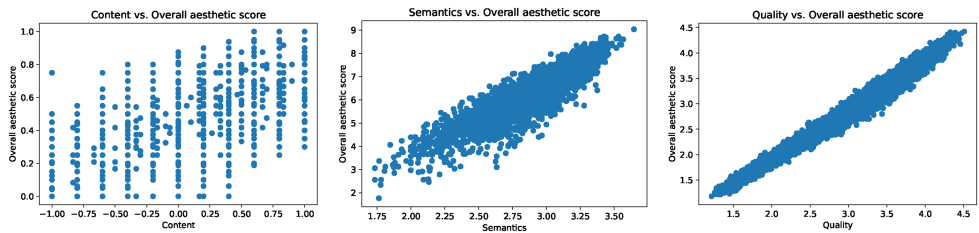

**FIGURE 11** Example scatter plots for each dataset, illustrating the relationship between the scores for one attribute (independent variable) and the overall aesthetic scores derived from the training data. (Left) A plot from the AADB dataset depicting the relationship between content and overall aesthetic score; (Middle) a plot from the EVA dataset illustrating the relationship between semantics and overall aesthetic score; (Right) a plot from the PARA dataset demonstrating the relationship between quality and overall aesthetic score.

## DISCUSSION

We have introduced alternative ML models and implemented SHAP analysis in the field of empirical aesthetics. A natural question that arises is why we chose these ML models instead of linear regression (LR), which is regularly implemented by psychologists. To address this, we first present the results of LR for all three datasets alongside a comparison with the SVR results, our best-performing ML model, in Table 5. As observed, LR demonstrates good performance in terms of all metrics across the three benchmarks.

However, another question arises at this point: *Are these datasets suitable for LR?* We must verify that the datasets meet the required assumptions for LR before its application. Checking the assumptions of an LR model is essential to ensure the validity and reliability of the results. If these assumptions are validated, applying LR becomes feasible; otherwise, the results may not be reliable. In the field of psychology, there is a common tendency among researchers to overlook this crucial step of validating LR assumptions, which can potentially lead to incorrect conclusions.

The fundamental assumption of LR is *linearity*, which indicates that the relationship between the independent variables (predictors) and the dependent variable (target variable) is *linear*. In Figure 11, we present a scatter plot as an example for each dataset. For the AADB dataset, the relationship does not appear to be clearly linear. As this pattern holds for most features (see Appendix G: Figure G1), LR might not be a suitable model for this dataset. However, the scatter plots for both the EVA and PARA datasets seem to support the linearity assumption (see Appendices H and I; Figures H1 and I1). The reason the LR model appears to be a good fit for the AADB dataset in Table 5 could be due to the true nature of the data or its representation in normalized form in this dataset. Sometimes, the underlying relationship between variables might be fundamentally linear, even if the data itself appears non-linear. Nonetheless, it is crucial to examine not only linearity but also other LR assumptions, such as homoscedasticity, normality, and multicollinearity, for the proper application and interpretation of LR models (Montgomery et al., 2021). Should significant deviations from these assumptions be found, alternative modelling strategies may need to be considered, or data transformation may be required. Regardless of the approach taken, whether it involves various models such as mixed-effects linear regression or ridge regression, it is essential to validate these assumptions.

What happens if we interpret the LR results for the AADB dataset, despite its not meeting the linearity assumption? Let's consider the LR equation for the AADB dataset as given in Equation 2.[4] According to this equation, colour harmony has the largest coefficient, suggesting that it is the most influential feature on the overall aesthetic score, followed by motion blur, depth of field, and content. However, the regression coefficients are very small. Conversely, in Figure 5, the SHAP summary plot indicates that content is the feature with the greatest importance on the model's outcome, which is the overall aesthetic score. This finding is supported by other ML models. Meanwhile, the LR coefficients suggest a different feature ranking, raising the question: *Which model would be better to interpret?*

$$\begin{aligned}\hat{y} =\ & -0.4599 + 0.1638 X_{BE} + 0.2760 X_{CH} + 0.2106 X_{CO} + 0.2142 X_{DoF} \\ & + 0.1896 X_{Light} + 0.2459 X_{MB} + 0.1482 X_{OE} + 0.0934 X_{Rep} + 0.1749 X_{RoT} \\ & + 0.1035 X_{Sym} + 0.1805 X_{VC}\end{aligned} \quad (2)$$

The answer is clear: Given that the AADB dataset does not meet the assumptions required for the LR model, a different regression model is more appropriate. Consequently, the results obtained from our ML models, particularly the best-performing SVR model, are more suitable in this case, consistent with findings from other ML models. Furthermore, we present the Spearman's rank correlations between the ground-truth overall aesthetic scores and attribute scores across all three datasets in Figure 12. Specifically, for the AADB dataset, content demonstrates the highest correlation with the overall score, a finding that is consistent with the SHAP results depicted in Figure 5. However, this contrasts with the LR results in Equation 2. In the case of the EVA dataset, composition and depth, as well as semantics, are the top two attributes, both showing very close and strong correlations with the overall score. These findings also align with our SHAP analysis. As for the PARA dataset, quality has the highest correlation with the overall aesthetic score. It is worth noting that object emphasis appears to have a low correlation with the overall score in the PARA dataset, potentially due to the attribute being represented by a binary label (see Section Datasets). Additionally, when we compare the attribute rankings in terms of their correlations with overall scores across all datasets in Figure 12, we observe a clear alignment with the SHAP analysis results seen in Figures 5, 7, and 9.

Another point we should emphasize pertains to the question: *What is the advantage of SHAP analysis?* LR coefficients indicate that a change in a predictor leads to a proportional change in the target variable. However, these coefficients do not directly inform us about *the overall importance* of a feature. While LR coefficients show what happens when we change the values of an input feature, they do not directly measure a feature's overall importance. This is because the value of each coefficient depends on the scale of the input features.[5] In contrast, SHAP provides global interpretations, revealing the importance of each feature across the entire dataset. Additionally, interpreting SHAP values offers a significant advantage for complex models like deep neural networks, where directly interpreting parameters is challenging due to the lack of simple coefficients.

*When would it be better to consider traditional models like LR?* LR is a suitable option if the dataset meets the model's necessary assumptions. Such models are advantageous for interpretability, allowing direct observation of how changes in features affect the outcome. The regression equations for the EVA and PARA datasets are presented in Equations 3 [6] and 4,[7] respectively. Upon examining the LR coefficients, we see that the results align with the SHAP analysis presented in Figure 7 and Figure 9, respectively. Moreover, we observe that the most important feature on predicting aesthetics is content in the AADB dataset, semantics on the EVA dataset, and quality in the PARA dataset. These results show that AI benchmark datasets do focus more on content rather than aesthetics. This observation empirically shows a gap in the literature and highlights the importance of future studies aimed at collecting more psychologically meaningful benchmarks.

---

[4]BE: balancing elements; CH: colour harmony; CO: content; DoF: depth of field; MB: motion blur; OE: object emphasis; Rep: repetition; RoT: rule of thirds; Sym: symmetry; VC: vivid colour.

[5]https://shap.readthedocs.io/en/latest/example_notebooks/overviews/An%20introduction%20to%20explainable%20AI%20with%20Shapley%20values.html.

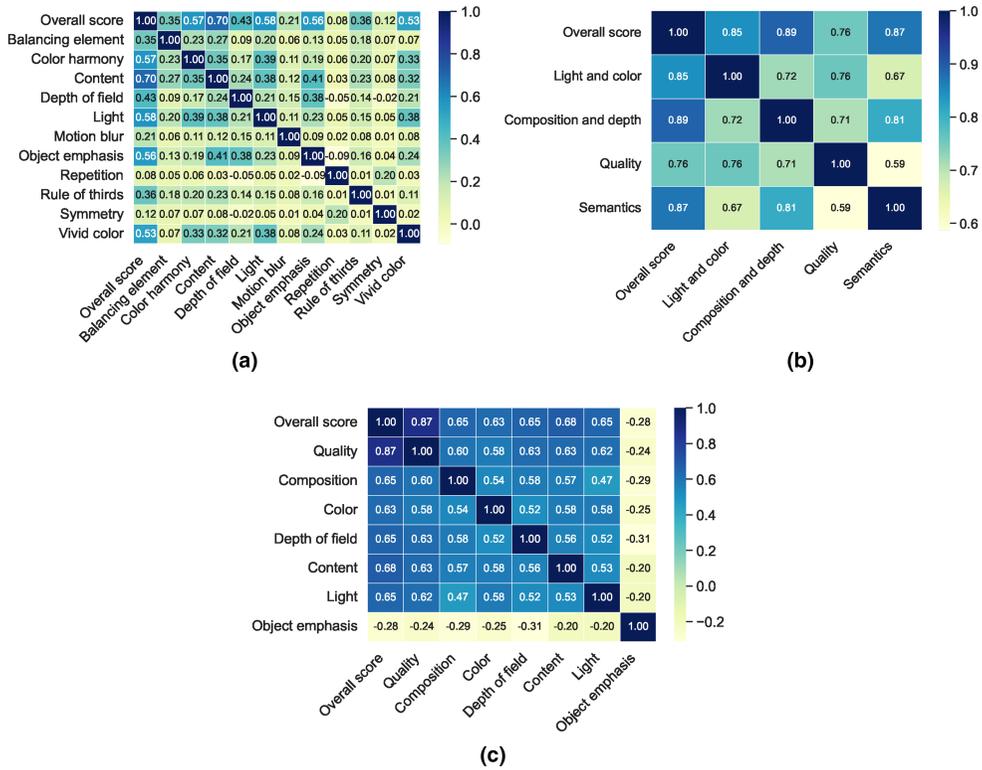

**FIGURE 12** Spearman's rank correlations between the ground-truth overall aesthetic scores and attribute scores across all datasets. (a) The AADB dataset; (b) The EVA dataset; (c) The PARA dataset.

$$\hat{y} = -0.0331 + 0.2671 X_{LC} + 0.2937 X_{Comp} + 0.1242 X_{Quality} + 0.3423 X_{Sem} \qquad (3)$$

$$\hat{y} = -0.0468 + 0.6017 X_{Quality} + 0.1171 X_{Comp} + 0.0946 X_{Color} + 0.0761 X_{DoF} + 0.0494 X_{Light} + 0.1315 X_{Content} - 0.0003 X_{OE} \qquad (4)$$

Notably, in our SVR models, we initially used the RBF kernel to demonstrate the model's ability in handling both linear and nonlinear data. Given the complexity of the decision function in an RBF kernel, it is challenging to derive straightforward linear coefficients or slopes. However, we modified the kernel to linear for comparative analysis and examined the coefficients. With the linear kernel, the SVR model achieved $R^2$ scores of .8684 for the AADB dataset, .9275 for the EVA dataset, and .9860 for the PARA dataset. The feature rankings from these SVR models (with a linear kernel) precisely align with our original results. The coefficients for the AADB dataset differ from the SHAP results, whereas for the other two datasets, they align with the SHAP findings.

---

[6] Comp: composition and depth; LC: light and colour; Sem: semantics.
[7] Comp: composition; DoF: depth of field; OE: object emphasis.

*When would it be better to consider alternative ML models?* There are several reasons to opt for alternative ML models. (1) *No strict assumptions*: The models presented in Section Machine learning models for regression can be directly applied and are capable of capturing patterns in data without the strict assumptions required for LR. (2) *Handling non-linearity*: In cases where data exhibits non-linear relationships, the presented ML models are effective in capturing these data, unlike LR. (3) *Handling feature interactions*: ML models are capable of automatically capturing interactions between features, a significant advantage over LR where explicitly creating those interaction terms is often necessary. This feature allows ML models to more effectively model complex relationships within the data.

Overall, applying LR is acceptable if the data at hand validate the assumptions. It is a traditional statistical technique that has been widely implemented in many fields, including psychology. On the other hand, alternative ML models offer greater flexibility and more straightforward approaches. These models do not require the validation of assumptions and can handle both linear and nonlinear data. Among the four ML models presented in this paper, only the MLP requires extensive hyperparameter tuning, in contrast to the other three models. However, except for the MLP, the remaining models are straightforward to implement. Moreover, SHAP analysis represents a powerful XAI technique, enhancing the interpretability of the trained models. The differences in interpretation between LR coefficients and SHAP values are particularly interesting. For more reliable and consistent results, we recommend not relying solely on a single model. Instead, employing multiple approaches and comparing their results is advisable to better understand the trade-offs involved in this kind of research. We propose the approach of applying various regression models to ensure both consistency and reliability in the findings.

# CONCLUSION

In this study, we have undertaken a new perspective of image aesthetics preferences by focusing on attribute scores and utilizing various ML models to predict overall aesthetic scores across benchmarks. Our emphasis on XAI, particularly the SHAP method, has allowed us to gain valuable insights into the contributions of individual attributes to the model's predictions. We observe the factors influencing image aesthetics and shed light on the interpretability and explainability of our ML models. Moreover, we apply the SHAP method in the field of computational aesthetics for the first time, also allowing us to examine the interactions between the features.

It is essential to acknowledge the inherent subjectivity of aesthetic preferences. Therefore, the quality and diversity of the dataset play a critical role in the performance and interpretability of our models. In this study, we utilize three image aesthetic assessment benchmarks, each with its own set of attributes and ratings per image. As a result, our data-dependent models may yield slightly different results. The importance of having a well-curated and diverse dataset cannot be understated, as it significantly impacts the generalization and robustness of findings.

Furthermore, examining various ML models enhances the consistency of results. Through this novel approach for aesthetics research and the application of XAI technique, we hope to enhance our understanding of aesthetics in images, contributing to the field of computational aesthetics. We emphasize that the model choice, as well as the utilization of SHAP, depends on the nature of the data, the importance of interpretability, and the need for predictive performance. Ultimately, our efforts aim to enhance the overall understanding and appreciation of image aesthetics through computational methods.

## AUTHOR CONTRIBUTIONS
**Derya Soydaner:** Conceptualization; methodology; formal analysis; software; writing – original draft. **Johan Wagemans:** Conceptualization; supervision; funding acquisition; writing – review and editing.

## ACKNOWLEDGEMENTS
Funded by the European Union (ERC AdG, GRAPPA, 101053925, awarded to Johan Wagemans). Views and opinions expressed are however those of the authors only and do not necessarily reflect those of the European Union or the European Research Council Executive Agency. Neither the European Union nor the granting authority can be held responsible for them.

## CONFLICT OF INTEREST STATEMENT
None declared.

## DATA AVAILABILITY STATEMENT
The data that support the findings of this study are available from the corresponding author upon request.

## ORCID
*Derya Soydaner*  https://orcid.org/0000-0002-3212-6711

# APPENDIX A

## Attributes in the AADB dataset

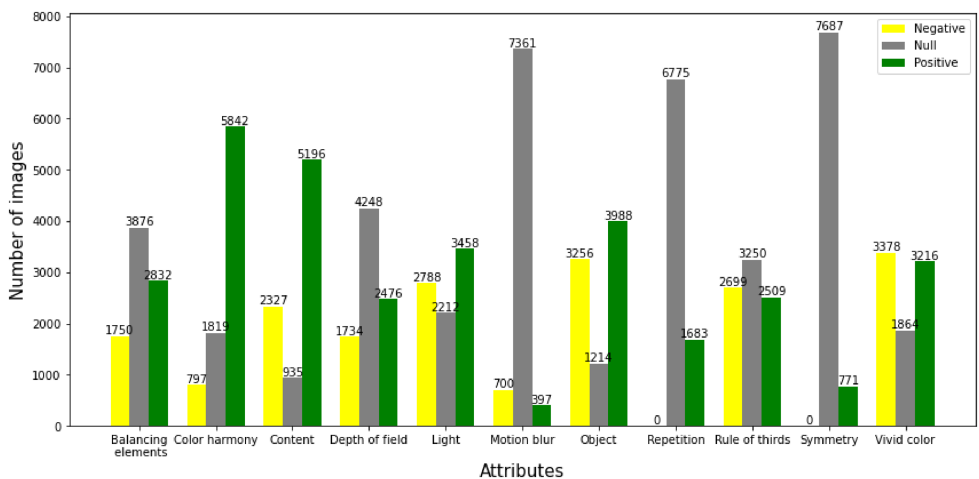

**FIGURE A1** Illustration of negative, null, and positive ratings for each attribute in the training set of AADB dataset (Kong et al., 2016; Soydaner & Wagemans, 2023).

# APPENDIX B

## SHAP summary plots for Random Forest, XGBoost, MLP, and Linear Regression for the AADB dataset

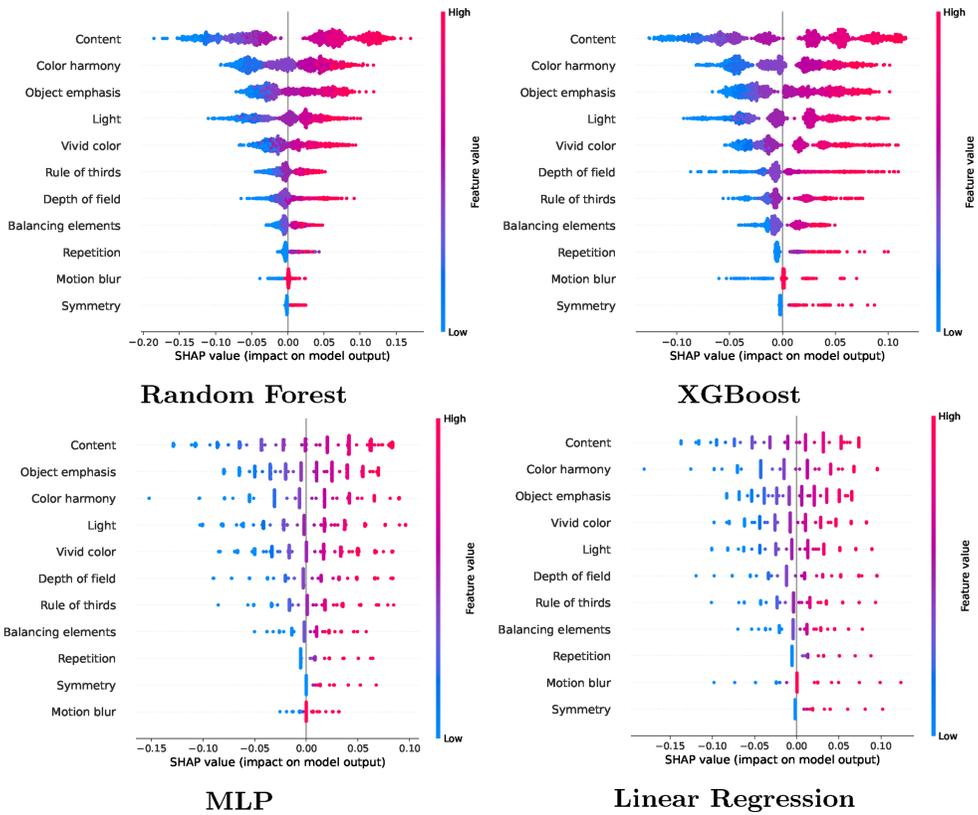

**FIGURE B1** SHAP summary plots for the AADB dataset based on the Random Forest, XGBoost, MLP, and linear regression models.

# APPENDIX C

## Interaction plots for the AADB dataset

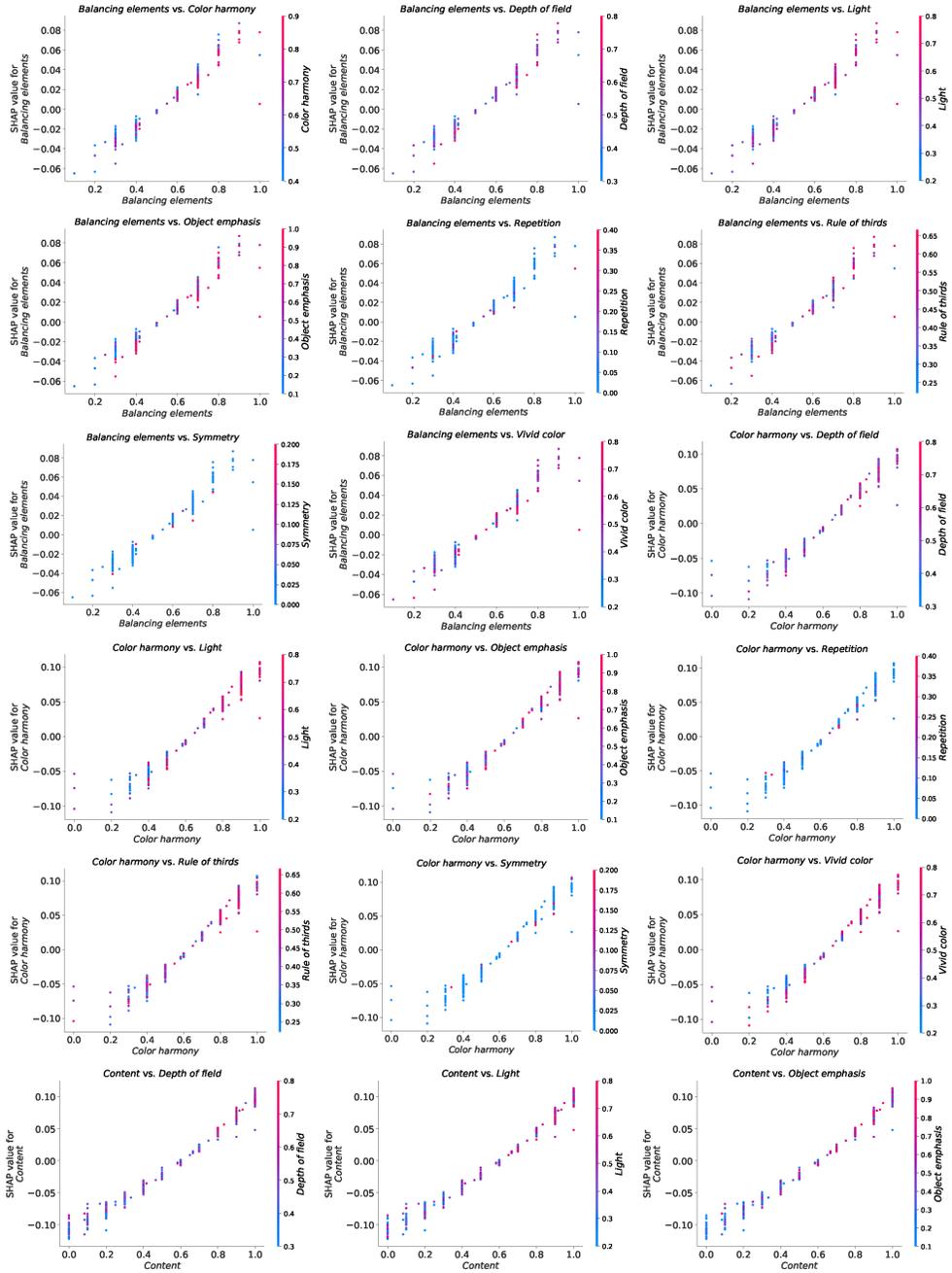

**FIGURE C1** Interaction plots for the AADB dataset based on the SVR model (part I).

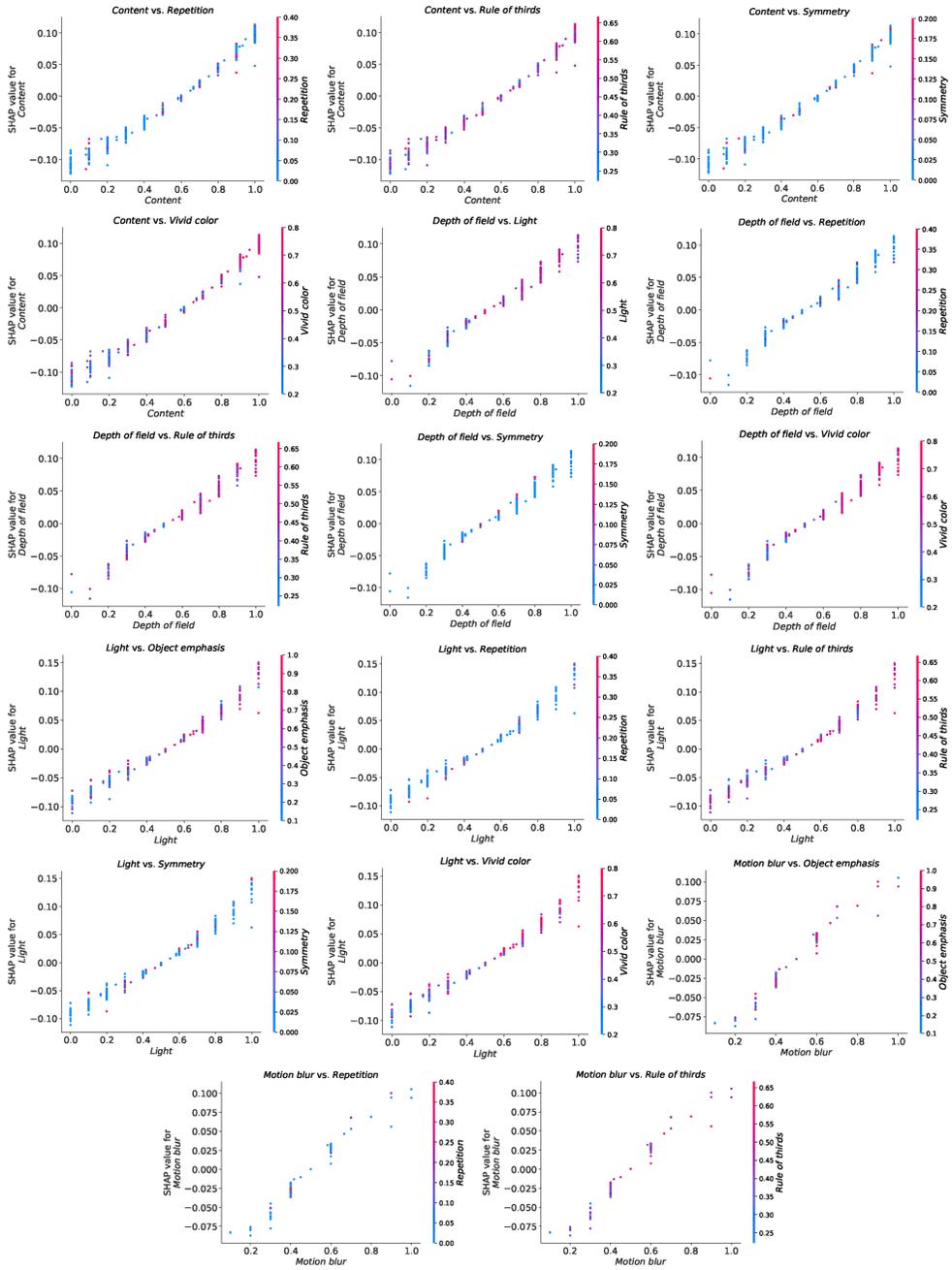

**FIGURE C2** Interaction plots for the AADB dataset based on the SVR model (part II).

**30** | SOYDANER and WAGEMANS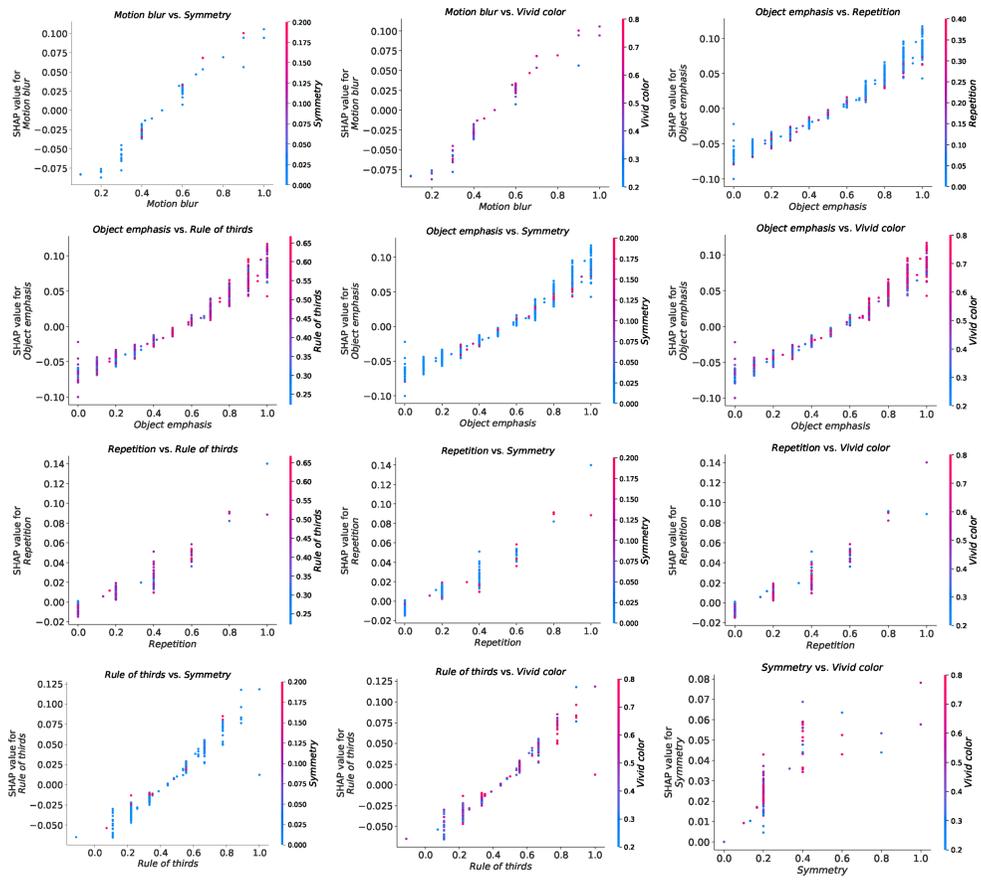

**FIGURE C3** Interaction plots for the AADB dataset based on the SVR model (part III).

## APPENDIX D

## SHAP summary plots for Random Forest, XGBoost, MLP, and Linear Regression for the EVA dataset

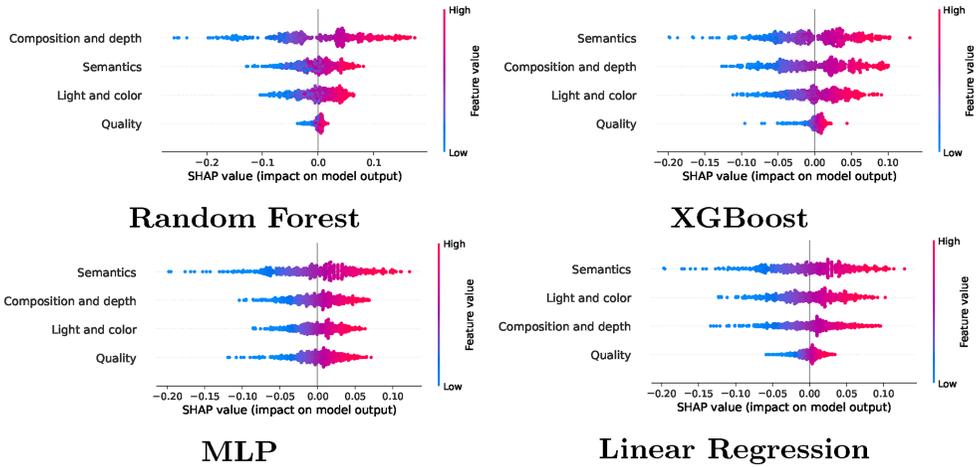

**FIGURE D1** SHAP summary plots for the EVA dataset based on the Random Forest, XGBoost, MLP and linear regression models.

## APPENDIX E

## SHAP summary plots for Random Forest, XGBoost, MLP, and Linear Regression for the PARA dataset

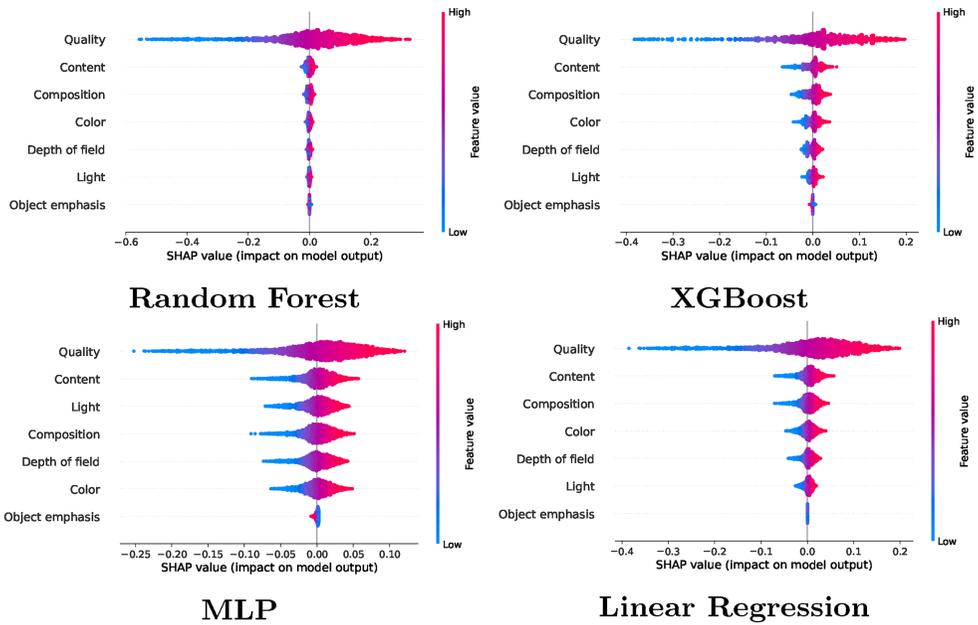

**FIGURE E1** SHAP summary plots for the PARA dataset based on the Random Forest, XGBoost, MLP and linear regression models.

# APPENDIX F

## Interaction plots for the PARA dataset

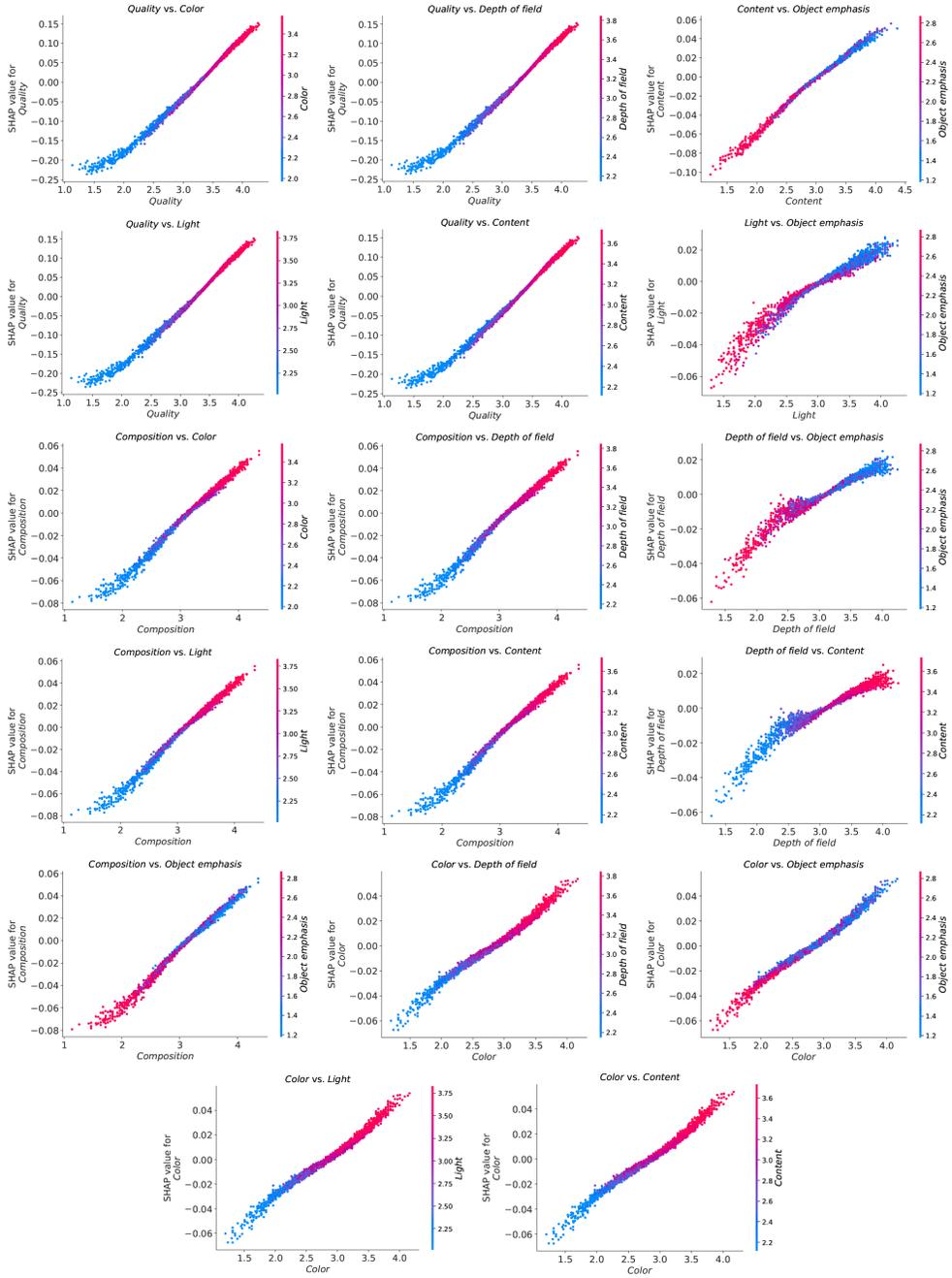

**FIGURE F1** Interaction plots for the PARA dataset based on the SVR model.

# APPENDIX G

## Scatter plots of input features versus target variable for the AADB dataset

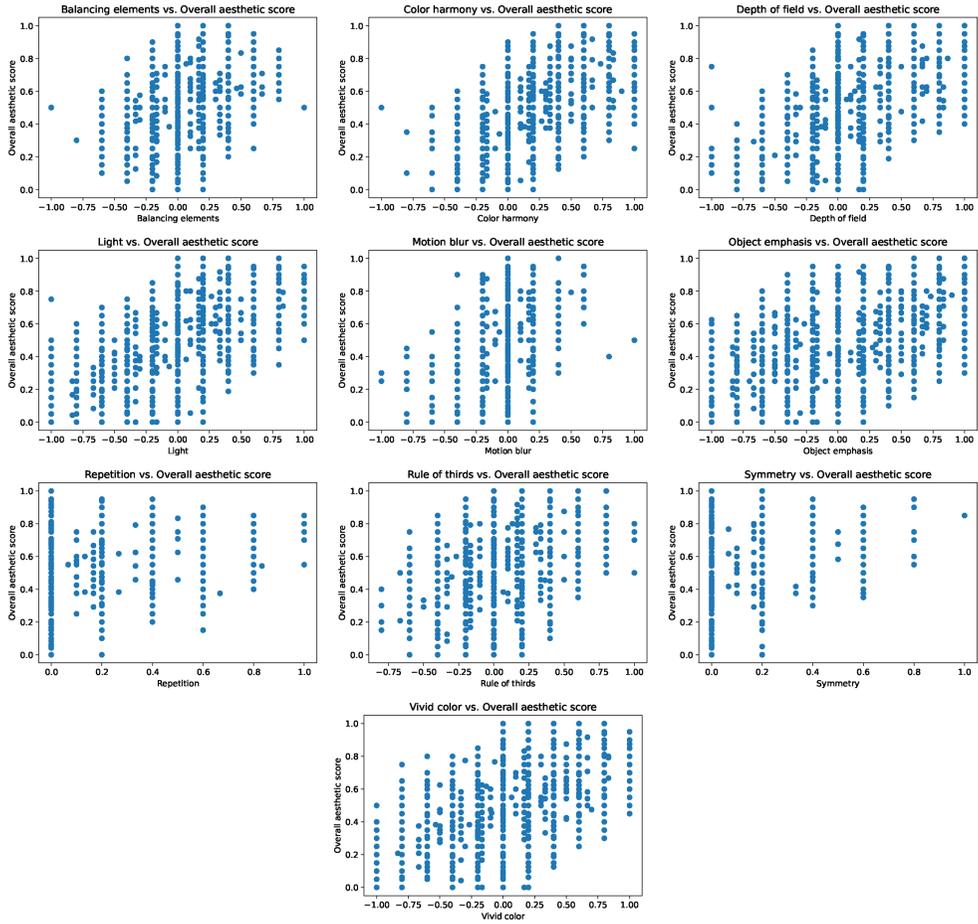

**FIGURE G1** Scatter plots for the AADB dataset.

# APPENDIX H

**Scatter plots of input features versus target variable for the EVA dataset**

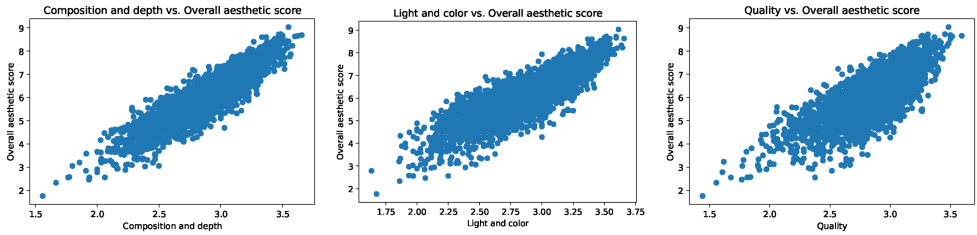

**FIGURE H1** Scatter plots for the EVA dataset.

# APPENDIX I

**Scatter plots of input features versus target variable for the PARA dataset**

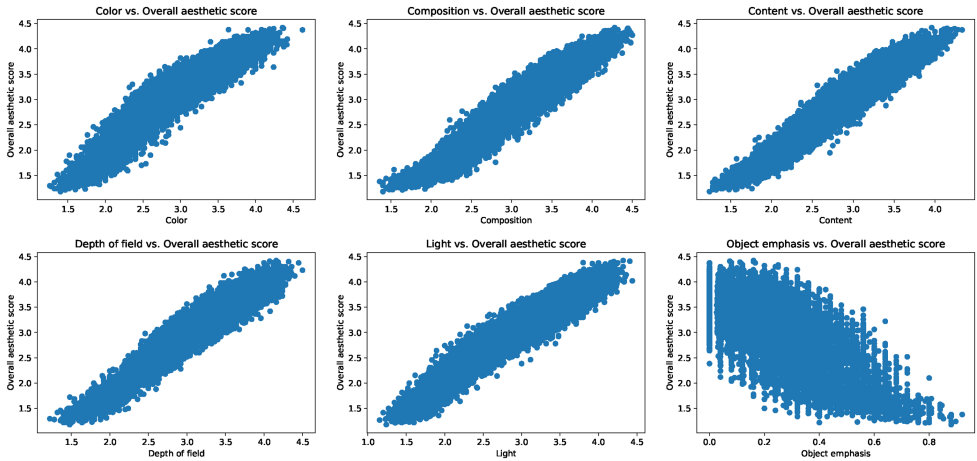

**FIGURE I1** Scatter plots for the PARA dataset.